\documentclass[ijds]{style_ms/informs4}
\OneAndAHalfSpacedXI
\usepackage{colortbl}
\usepackage{pifont}

\usepackage{listings}
\usepackage{lstfiracode}

\usepackage{amsmath,amssymb,amsfonts}
\usepackage{booktabs}
\usepackage{tcolorbox}
\usepackage{enumitem}
\usepackage{hyperref}
\usepackage[capitalise]{cleveref}

\usepackage{natbib}
 \bibpunct[, ]{(}{)}{,}{a}{}{,}%
 \def\bibfont{\small}%
\usepackage{pgfplots}
\pgfplotsset{compat=1.18}

\definecolor{OGreen}{RGB}{220, 230, 214}
\definecolor{OYellow}{RGB}{249, 242, 209}
\definecolor{OGray}{RGB}{245, 245, 245}

\usepackage{rotating}
\usepackage{fancyvrb}

\definecolor{main1}{HTML}{6C8EBF}
\definecolor{main2}{HTML}{B85450}
\definecolor{main3}{HTML}{82B366}

\definecolor{BGcolor}{HTML}{F5F5F5}
\definecolor{light1}{HTML}{FFF2CC}
\definecolor{light2}{HTML}{D5E8D4}

\usepackage{algorithmicx}
\usepackage{algorithm}
\usepackage{algpseudocode}

\usepackage{tabularx}
\usepackage{longtable}
\usepackage{booktabs}
\usepackage{array}

\usepackage{footnotehyper}
\makesavenoteenv{table}
\makesavenoteenv{tabular}

\definecolor{deeppink}{RGB}{255,20,147}

\newcommand{\cledit}[1]{\textcolor{black}{#1}}
\newcommand{\caledit}[1]{\textcolor{black}{#1}}

\newcommand{\tmop}[1]{\ensuremath{\operatorname{#1}}}

\RequirePackage{tgtermes}
\RequirePackage{newtxmath}
\RequirePackage{bm}
\RequirePackage{endnotes}

\usepackage{natbib}
 \bibpunct[, ]{(}{)}{,}{a}{}{,}%
 \def\bibfont{\small}%
\EquationsNumberedThrough    

\TheoremsNumberedThrough     
\ECRepeatTheorems  %

\MANUSCRIPTNO{MNSC-0001-2024.00}

\begin{document}
\RUNAUTHOR{AhmadiTeshnizi et al.}
\RUNTITLE{OptiMUS-0.3}

\TITLE{OptiMUS-0.3: Using large language models to model and solve optimization problems at scale}

\ARTICLEAUTHORS{%
\AUTHOR{Ali AhmadiTeshnizi\textsuperscript{a} , Wenzhi Gao\textsuperscript{b}, Herman Brunborg\textsuperscript{b} Shayan Talaei\textsuperscript{a}, Connor Lawless\textsuperscript{a}, Madeleine Udell\textsuperscript{a,b}}
\AFF{\textsuperscript{a} School of Management Science and Engineering, Stanford University, Stanford, California 94305}
\AFF{\textsuperscript{b} Institute for Computational and Mathematical Engineering, Stanford University, Stanford, California 94305}
}

\ABSTRACT{
Optimization problems are pervasive in sectors from manufacturing and distribution to healthcare.
However, most such problems are still solved heuristically by hand rather than optimally by state-of-the-art solvers because the expertise required to formulate and solve these problems limits the widespread adoption of optimization tools and techniques. We introduce a Large Language Model (LLM)-based system designed to formulate and solve (mixed integer) linear programming problems from their natural language descriptions. Our system can develop mathematical models, write and debug solver code, evaluate the generated solutions, and improve efficiency and correctness of its model and code based on these evaluations. \caledit{OptiMUS is designed as a productivity tool for optimization practitioners who understand the problem domain and can describe it precisely, but seek to accelerate the modeling and implementation workflow.} OptiMUS-0.3 utilizes a modular structure to process problems, allowing it to handle problems with long descriptions and complex data without long prompts. 
\caledit{Experiments demonstrate that OptiMUS-0.3 outperforms direct-prompting baselines by over 43\% on easy and 18\% on hard instances. It remains competitive with fine-tuned specialist models on benchmark problems, and outperforms them on real-world case studies (28.6\% vs.\ 0\%) where fine-tuned models fail to generalize. }
Ablation studies show that modular architecture with error correction is central to these gains. 
\caledit{A key finding is that system architecture is a stronger driver of performance than model capability. Structured decomposition with targeted error correction enables weaker models to match stronger models under naive prompting, and remains competitive with fine-tuned specialist models without retraining costs.}

}


\maketitle

\section{Introduction}
\label{intro}

Optimization problems are ubiquitous in domains from operations and economics to engineering and computer science. While major advances in optimization algorithms over the last several decades have led to reliable and efficient optimization methods for a wide variety of structured optimization problems, including linear programming (LP) and mixed-integer linear programming (MILP), optimization modeling --- transforming a domain problem into a mathematical optimization problem --- still requires expert knowledge. According to a recent survey, 81\% of Gurobi's commercial solver users have advanced degrees, with 49\% of them holding a degree in operations research \citep{gurobiReport}.
This expertise gap prevents many organizations from using optimization, even when it could significantly improve their operations. Examples include inventory management in supermarkets, patient operations in hospitals, transportation policies in small municipalities, energy management in local solar farms, and operations in small businesses or NGOs \citep{saghafian2015operationsHospital, aastrup2010fortyRetail, yao2020optimizationTransport, shakoor2016wakeFarm}. 
Automating optimization modeling would allow optimization practitioners to tackle more problems and develop models faster, amplifying the impact of existing expertise.

In this paper, we develop an automated system to accelerate the workflow of existing optimization practitioners. We specifically focus on helping optimization users develop, tune, and implement (mixed-integer) linear optimization problems, a flexible and broad class of optimization problems that have been applied successfully in a wide range of real-world use cases such as scheduling and production planning. Our work aims to improve the productivity of optimization practitioners---for example, an MBA analyst in industry who understands the problem domain but wants to model and implement faster---akin to the productivity gains realized by software engineers with systems such as GitHub CoPilot. \caledit{This work is a step towards systems that could ultimately broaden access to optimization tools for users with less formal training, but that goal is beyond the scope of the current paper.} Given that our focus is on existing optimization practitioners, we assume for the remainder of this paper that the user has already identified a problem for which MILP is an effective tool and can describe the problem in a level of detail such that modeling is possible. Aiding users to find the correct modeling paradigm (e.g., constraint programming, conic optimization) and supporting them in concretizing vague problems remains an important open problem for future research.

Large language models (LLMs) offer a promising way to automate optimization modeling, and increase the reach of the powerful algorithms developed by the operations research community. 
\caledit{However, using LLMs for optimization modeling presents unique challenges that arise from the scale and complexity of real-world optimization problems. Scalability in optimization modeling arises along several distinct dimensions: (i) \emph{description length}, the length of the natural-language problem description; (ii) \emph{problem complexity}, the number and type of decision variables, constraints, and logical or combinatorial structures (e.g., binary variables, special ordered sets, logical constraints); and (iii) \emph{data scale}, the size of the structured input data (e.g., number of customers, products, or time periods). In this work, we provide empirical evidence of scalability with respect to description length and problem complexity, and present initial evidence on data scale. Beyond these scalability challenges, LLMs still suffer from major flaws that prevent deployment in important applications.}
We highlight four challenges:

\begin{itemize}[leftmargin=10pt]
    \item \textbf{Long problem descriptions.}
    Realistic optimization problems can be exceedingly long and complex: for example, the documentation for the energy system problem described
    in \cite{Challenge3} is 60 pages long.
    Unfortunately, LLMs have a limited context size, and even long-context models perform less well as the input context grows \citep{liu2023lostinthemiddle, LengthReasoning}. 
    Consequently, LLMs tend to make more mistakes as the length of the problem description increases and perform poorly on complex problems. 
     \item \textbf{Large problem data.} 
     \caledit{Optimization models frequently depend on structured input data such as customer attributes, network structures, or time-indexed parameters. As the size of these datasets grows, modeling systems must correctly incorporate this data into the formulation without exceeding prompt limits or introducing modeling errors.}
     
    \item \textbf{Hallucination.} 
    LLMs are known to \emph{hallucinate}: they may produce answers that sound reasonable, but are incorrect. 
    In the context of optimization, the generated solver code may hallucinate constraints that incorrectly model the problem, 
    or it may hallucinate API calls that do not exist, resulting in code that cannot run.
    It is especially challenging to verify whether the solution is correct, supposing the code runs without error. 
    For instance, if the solver claims the solution is unbounded, 
    perhaps a constraint has been accidentally omitted from the formulation.

    \item \textbf{Bad models.} 
    The solve time for an optimization problem can depend on the particular modeling formulation chosen, and on how the structure of the problem is communicated to the solver. Optimization experts spend much of their effort modeling the problem to enhance the efficiency of the solution method. 
    A challenge for LLMs is to produce not just an accurate model, 
    but good code that solves the problem quickly.
\end{itemize}

This paper studies the feasibility of using LLMs to expand the reach of operations research techniques and analytic methods for decision making by studying the performance of OptiMUS, an automated system we have developed that uses LLMs combined with traditional solvers to model and solve (mixed integer) linear programs (MILPs). Our studies of OptiMUS demonstrate both the promise and pitfalls of the LLM approach to quantitative decision making,
and help clarify the effectiveness of LLMs for optimization modeling as of 2026. More broadly, through ablation studies, this paper shows how structured reasoning pipelines provide a clear pathway to address current limitations and enhance the reliability of LLM-based systems in optimization. Through systematic ablation studies, we find that the primary bottleneck in LLM-assisted optimization modeling is system architecture, not LLM capability. Modular decomposition with targeted error correction enables weaker models to match stronger models using naive prompting, and matches fine-tuned specialist models without retraining costs. 

\caledit{We organize our evaluation of scalability around the three dimensions introduced above.
\begin{itemize}[leftmargin=15pt]
    \item \textbf{Description length.} We demonstrate scalability with respect to description length through the NLP4LP benchmark (\cref{table:datasets}). Problems in the hard subset have descriptions averaging 912 characters (nearly twice the length of the easy subset), and the case study instances average 2,578 characters, an order of magnitude longer than textbook instances. Performance on these subsets provides direct evidence of the system's ability to handle increasingly long problem descriptions.
    \item \textbf{Problem complexity.} We evaluate scalability with respect to problem complexity through the seven case-study instances (\cref{sec:dataset}), which are drawn from published real-world applications and involve rich combinatorial structure: multiple classes of integer and binary variables, logical constraints, and domain-specific modeling patterns that differ substantially from textbook problems.
    \item \textbf{Data scale.} OptiMUS handles large input data by separating problem data from the natural-language description: data is stored externally and referenced symbolically in the model, so prompt length does not grow with data size. We validate this empirically in \cref{sec:large-scale}: comparing OptiMUS against a naive prompting baseline (full data provided in prompt) across input sizes from 3\,KB to 935\,KB, OptiMUS maintains 100\% success rate throughout while naive prompting degrades and fails entirely once data exceeds GPT-4o's context window.
\end{itemize}
We do not make claims about scalability along other dimensions, such as solver-level scalability (handling models with thousands of variables or constraints). These are important directions for future work but are outside the scope of the current empirical evaluation.}

Our contributions in this paper are as follows:
\begin{itemize}[leftmargin=15pt]
    \item \caledit{We curate NLP4LP, a comprehensive open-source dataset of 361 optimization problems organized into three subsets: an \emph{easy} subset of 289 LP problems with short descriptions and scalar parameters; a \emph{hard} subset of 65 LP/MILP problems with longer descriptions and multi-dimensional parameters; and a \emph{case study} subset of 7 real-world MILP problems drawn from the INFORMS Journal on Applied Analytics, featuring complex combinatorial structure and large-scale input data. \cref{table:datasets} compares NLP4LP to existing datasets and \cref{sec:dataset} describes NLP4LP in detail. Our dataset also includes problems that are less likely to have been used to train existing LLMs, reducing the risk of data leakage compared to previous optimization modeling datasets.} 
    
    \item We develop a modular, LLM-based agent to model and solve optimization problems, which we call OptiMUS (or, occasionally, OptiMUS-0.3 to distinguish it from prior versions). 
    OptiMUS-0.3 employs a connection graph that allows it to process each constraint and objective independently,
    which \cledit{allows it to process problems with long natural-language descriptions and large input datasets without requiring excessively long prompts.} 
    
    \item We develop several modules designed to improve the performance of OptiMUS and study their impact, including:
    \begin{enumerate}[leftmargin=15pt]
    \item \emph{Self-reflective error correction.} We ask OptiMUS to evaluate and correct its output, and assess its own confidence in its output, allowing it to fall back to a more powerful LLM or to user feedback when OptiMUS is unsure its answer is correct.
    \item \emph{Advanced optimization modeling techniques.} We teach OptiMUS advanced optimization modeling techniques by prompting it to identify important structures in the problem, including structure of problems and solver features such as special ordered sets, and to use those structures to model and solve the problem more efficiently. 

    \end{enumerate}

    \item We perform ablation studies of the components of our framework to demonstrate which elements improve performance and which can actually degrade performance. We find OptiMUS-0.3 beats GPT-4o alone by over 43\% on easy instances and over 18\% on hard instances. \caledit{On seven real-world case-study instances, OptiMUS-0.3 (GPT-4o) solves 28.6\% of instances, compared to 14.3\% for direct prompting with GPT-4o and 0\% for fine-tuned specialist models.}

    \item 
    \cledit{We develop a publicly available web application that allows practitioners to try OptiMUS. This webapp enables human-in-the-loop automated optimization modeling: users can edit LLM outputs and OptiMUS can flag low-confidence LLM outputs to solicit corrections from the user.} 
    

\end{itemize}

An initial version of this work was published in a conference proceeding \citep{optimus-0.2}, that introduced a smaller version of the NLP4LP dataset, and a less robust LLM agent-based system for modeling optimization problems that we will refer to as OptiMUS-0.2. The new enlarged version of the NLP4LP dataset which accompanies this paper has been open-sourced and includes a modular design that enables future researchers to improve on different sub-problems within our framework such as parameter extraction, modeling, or coding. Compared to the earlier version, the modular design of OptiMUS-0.3 substantially improves the robustness of the system for more complicated modeling problems. 
New modules that enable the system to leverage advanced solver functionality, such as special-order-set constraints, can dramatically speed up the solution time for the associated optimization problems.
Additional LLM techniques in OptiMUS-0.3 such as self-reflective error correction improve performance. 
Finally, this paper is the first to present the OptiMUS webapp.

\section{Related Work}

\label{background}

Formulating an optimization model is often a challenging task even for experts in optimization.
Different formulations can lead to significantly different solving times and
enable the use of different solvers or solution techniques \citep{BoydConvex}.
One important skill for an optimization expert is to identify assumptions or relaxations that
allow for casting the problem as a well-studied problem type, such as MILP, which
enables the use of well-developed optimization solvers.
Crafting such an efficient formulation often requires specialized knowledge \citep{ConicRelaxationSurvey, RelaxationOptimalFlow, roubivcek2020relaxation, SemidefiniteRelaxation, LagrangianRelaxationSurvey}. 

Given the formulation, an optimization expert must choose a solver. Each solver has a distinct interface and capabilities, with associated benefits and downsides \citep{achterberg2019gurobi, diamond2016cvxpy, manual1987ibmcplex}. However, the user manuals for these solvers are often hundreds of pages, making them difficult to understand and use. 

\paragraph{Progress in LLMs.} Recent progress in Natural Language Processing (NLP) has led to the development of LLMs useful for tasks such as answering questions, summarizing text, translating languages, and coding
\citep{openai2023gpt4, touvron2023llama, chowdhery2022palm, wei2023chainofthought,gao2023pal,borgeaud2022improving}. Connections to other software tools extend the reach and accuracy of LLMs, as demonstrated by plug-ins for code writing and execution \citep{paranjape2023art, wei2023chainofthought}.
\cite{yang2023LLMsAsOptimizers} use LLMs to directly generate solutions to optimization problems without calling traditional solvers through prompt optimization to improve performance.
The approach is limited to small problems since 
the performance of LLMs degrades as the input context grows, even for explicitly long-context models \citep{liu2023lostinthemiddle}. 

\paragraph{Chatbots for Optimization.} In a recent paper, \cite{chen2023diagnosing} developed a chatbot to help users detect and fix infeasible optimization problems expressed in \texttt{Pyomo} code and serves as an AI assistant rather than as a solver. \cite{li2023large} designed a chatbot to answer natural-language queries about a supply chain optimization model. \cite{mindoptCopilot} also developed a chatbot to facilitate optimization modeling, but there is no public paper or documentation available on it. \cite{lawless2024want} explored the use of LLMs to allow users to customize a simple constraint programming model in the context of meeting scheduling, but does not model more general MILP optimization problems.

\paragraph{Benchmark-driven Optimization Modeling.} 
More closely related to our approach, \cite{ramamonjison2023nl4opt} introduced a dataset of 1101 natural language representations of LP problems. They proposed a two-stage mapping from the natural-language representation to the problem formulation using an intermediate representation. \cite{ramamonjison-etal-2022-augmenting} designed a system to simplify and improve the modeling
experience for operations research, but did not offer an end-to-end solution. \cite{xiao2023chain} presented a multi-agent cooperative framework called Chain of Experts (CoE) to automatically model and program complex operation research (OR) problems, and evaluated it on NL4Opt and another more complex dataset, ComplexOR, introduced in that paper. \cite{astorga2024autoformulation} present a Monte-Carlo Tree Search (MCTS) based approach that decomposes the modeling problem into stages (e.g., modeling variables, then objectives, then constraints, etc.) and uses MCTS to explore the space of plausible models. As mentioned in the paper, this approach is not directly comparable to OptiMUS as it outputs a set of functionally distinct models as opposed to a single formulation and answer. Concurrent with the preparation of this paper, \cite{tang2024orlmtraininglargelanguage} introduced a new approach to OR modeling with LLMs that uses fine-tuning on a semi-synthetic dataset rather than prompt optimization or agentic models, which they call ORLM (Operations Research Language Model). 
\cite{yang2024optibench} also explore a synthetic data generation and fine-tuning approach for a broader class of optimization problems. These fine-tuning approaches are complementary to our work, as any progress towards more capable LLMs for optimization modeling can be combined with our modular prompting framework. 
Traditional MILP solvers generally benchmark against the \texttt{MIPLIB} benchmark \citep{gleixner2021miplib}, which offers a diverse collection of MILP problems in standard form. Unfortunately, most of these problems are not associated with a natural-language description, and so cannot be used to study optimization modeling as we do in this paper.

\section{Dataset}
\label{sec:dataset}

As part of this paper, we introduce a comprehensive open-source dataset of \caledit{361} optimization problems we call NLP4LP. The dataset builds upon earlier datasets for automated modeling, notably the NLP4OPT dataset \cite{ramamonjison2023nl4opt}, but includes a much richer set of information about the instances including an optimal solution and associated code.
Our goal in creating NLP4LP is to provide the community with examples they can use to design optimization modeling tools using natural language systems and to assess their quality. \caledit{The problems in NLP4LP are partitioned into three subsets: 1) an easy dataset that contains only LP problems with short descriptions and scalar parameters, 2) a hard dataset that contains both LP and MILP problems with longer descriptions and multi-dimensional parameters, and 3) a case study dataset drawn from real-world applications published in the INFORMS Journal on Applied Analytics (\cref{table:datasets}).} Unlike other modeling datasets, each instance in the NLP4LP dataset is accompanied by associated code to run the instance, and ground-truth intermediary representations of the problem (i.e., extracted parameters and targets, 
a list of clauses of the problem represented in natural language,
\LaTeX, and code, 
and a solution and optimal value for given problem data). 
This additional structure enables future researchers to investigate and improve different modules of the OptiMUS framework such as parameter extraction or code implementation, 
and even test these improvements with the OptiMUS webapp, 
without implementing the full software stack from scratch. 
Compared to the initial version of the NLP4LP dataset presented in \citet{optimus-0.2}, this version contains more instances from a wider breadth of domains and more complex problems from published papers, and includes additional information for each instance, such as code, as discussed earlier. The dataset is available on Hugging Face and in the supplementary material. 

Given the complexity of the task, and the components of each instance (description, sample feasible data, and solution code), gathering high-quality optimization instances is expensive. Unlike traditional ML datasets, NLP4LP does not provide a training set with the same distribution as the test set. As the cost of collecting each instance is high, we prioritize creating a large test set rather than reducing the size of the test set to provide training data. Consequently, the easy and hard subsets together comprise 354 problems, partitioned into 1) a development (dev) set of 23 problems (12 easy and 11 hard) and 2) a test set of 331 problems (277 easy and 54 hard). The dev set is intended to be used to develop optimization modeling tools, while the test set should be used to evaluate such tools. In our experiments and system design, we did not use any part of the test set for development, ensuring an unbiased evaluation of the models.

\caledit{The third subset consists of 7 case-study instances drawn from real-world operations research applications published in the INFORMS Journal on Applied Analytics (see \cref{table:case-studies}). Each instance was manually formulated and solved by the authors based on the corresponding published paper, with synthetic data generated to match the scale described therein. These problems are substantially longer and more complex than the easy and hard subsets, featuring multi-indexed parameters, rich domain-specific structure, and descriptions averaging 2{,}578 characters.}

\begin{table}[ht]
\centering
\caption{\caledit{Real-world case studies in the NLP4LP case study subset.}}
\label{table:case-studies}
\small
\begin{tabular}{clp{5.8cm}l}
\toprule
ID & Domain & Description & Reference \\
\midrule
355 & E-commerce & \textbf{Amazon Locker Capacity Management}: maximize locker throughput by reserving capacity per shipping option subject to capacity and demand constraints. & \cite{sethuraman2024amazon} \\
356/357 & Education & \textbf{Hybrid Class Scheduling}: two variants minimizing excess in-person attendance and maximizing student interaction, respectively, under social distancing capacity limits. & \cite{moallemi2024hybrid} \\
358 & Social Policy & \textbf{Human Trafficking Interception}: evaluate transit monitoring station efficiency via Data Envelopment Analysis. & \cite{dimas2023estimating} \\
359 & Urban Services & \textbf{Waste Collection Routing}: minimize total route length for a city-wide waste collection vehicle fleet. & \cite{bertero2023developing} \\
360 & Retail Logistics & \textbf{Rich Vehicle Routing}: minimize delivery cost for a DIY retailer using dedicated LSP tours and common carriers with a heterogeneous fleet. & \cite{tuma2024optimal} \\
361 & E-commerce & \textbf{Order Fulfillment Consolidation}: maximize multiorder hits using an LP-based pool management policy with machine-learned consolidation probabilities. & \cite{wang2024data} \\
\bottomrule
\end{tabular}
\end{table} 

Optimization modeling is used in many domains. To ensure that NLP4LP comprehends a wide variety of use cases, we first created a list of important types of optimization problems, such as scheduling, cutting, routing, blending, and packing, and a list of common application domains such as sports, government, retail, agriculture, and energy.
We tagged each problem NLP4LP with all relevant labels, and continued to gather instances until the dataset contains at least two instances per label. Instances in our dataset are drawn from several sources: some were created by our research team, inspired by problems in textbooks, lecture notes, and research papers, and some are drawn from existing datasets on LP modeling \citep{intro_to_opt, model_building, lectures_in_lp_modeling, xiao2023chain,ramamonjison2023nl4opt}. Most problems have around four labels, and some have as few as two or as many as eight.
For more information on problem types and domains refer to Appendix \ref{app:dataset}. \caledit{Real-world optimization problems are substantially longer and more complex than textbook instances. Table~\ref{table:datasets} shows that the NLP4LP Hard subset (avg.\ 912 characters) and Case Study subset (avg.\ 2{,}578 characters) contain problems with descriptions far longer than those in existing MILP modeling datasets, and that the case-study instances seriously challenge the modeling capabilities of current LLM systems (Section~\ref{sec:large-scale}).} 

One major concern in evaluation of LLM-based systems is data leakage: has the LLM been trained on the data in the purported test set? This issue is serious and has been known to contaminate the results of several important studies \citep{oren2023provingtestsetcontamination}.
Our goal is not to claim that the dataset is completely original (as it builds upon previous datasets), but rather ensure that all instances in our dataset have not been leaked to model pre-training. In our study, we take several measures to mitigate the risk of leakage, balancing this concern against the goal of broad coverage in our test set.
\begin{itemize}
\item Our dataset is guarded by Captcha, and only authenticated Hugging Face users can access it after agreeing to terms and conditions. These safeguards prevent the dataset from being used to train future LLMs.
\item Many problems in the dataset are inspired by textbook examples and recent publications, yet their text is entirely original. Moreover, the sources of these questions do not include solutions (either in \LaTeX~or code). We verified that the content of these questions does not appear anywhere on the internet using a plagiarism detection tool, which yielded a mean originality score of $84.9\%$ and a median originality score of $100\%$. See Appendix \ref{app:dataset} for details.
\item \caledit{The easy subset of NLP4LP is built directly on top of the NL4Opt dataset \citep{ramamonjison2023nl4opt}: the problem descriptions and formulations are the same, but each instance has been extended with the additional annotations required for our benchmark, namely associated solution code, ground-truth optimal values and solutions, extracted parameters, and clause-level natural language representations. This enrichment enables future researchers to evaluate and improve individual components of the modeling pipeline (e.g., parameter extraction, clause modeling, code generation) independently, which NL4Opt does not support. None of the MILPs, which represent the most challenging problems in our dataset, are sourced from existing datasets.} 
\end{itemize}

\begin{table*}[t]
\caption{A comparison on different aspects of complexity for various datasets. The unit for description length is characters}
\label{table:datasets}
\begin{center}
\begin{small}

\begin{tabular}{lccc}
\toprule
Dataset & Description Length & Instances (\#MILP)  & Multi-dimensional Parameters \\
\midrule
NL4Opt     & 518.0 $\pm$ 110.7 & 1101 (0) & $\times$ \\
ComplexOR & 497.1 $\pm$ 247.5 & 37 (12) & $\checkmark$\\
NLP4LP Easy (Ours) & 507.2 $\pm$ 102.6 & 289 (0) & $\checkmark$\\
NLP4LP Hard (Ours) & 912.3 $\pm$ 498.2 & 65 (18) & $\checkmark$\\
\caledit{NLP4LP Case Study (Ours)} & \caledit{2578.0 $\pm$ 976.5} & \caledit{7 (7)} & \caledit{$\checkmark$}\\
\bottomrule
\end{tabular}

\end{small}
\end{center}
\end{table*}

\section{Methodology} \label{sec:methodology}

An optimization problem proposes to maximize or minimize a given function $f$ over a set of allowable inputs ${\cal X}$. This paper focuses on modeling mixed-integer linear optimization problems (MILP) where $f$ is a linear function and the set of allowable inputs ${\cal X}$ can be represented by a set of linear inequalities over integer and real valued variables. A given MILP comprises three key components: (1) a set of $n$ discrete and real variables (i.e., the inputs that can be changed), (2) a set of linear inequalities and equalities that define different constraints on the input variables, and (3) a linear objective function that defines what the problem aims to minimize or maximize. A MILP can be written mathematically as
\begin{align*}
  \underset{\{ x \}}{\text{minimize}}\quad  & \sum_{j = 1}^n c_j x_j
  \\
  \text{subject to}\quad & \sum_{j = 1}^n a_{i j} x_j \leq b_i , \quad i = 1, \ldots m_1
  \\
  \quad & \sum_{j = 1}^n a_{i j} x_j = b_i , \quad i = 1, \ldots m_2 \\
  & x \in \mathbb{Z}^{n-r} \times \mathbb{R}^r.
\end{align*}

A feasible point $x^*$ that minimizes the objective function is called a (optimal) solution and has associated optimal objective value $c^T x^*$. For brevity, we refer to the equations defining constraints and the objective as different \textit{clauses} in the optimization model. A \textit{parameter} of an optimization model is any numeric coefficient in the optimization model (i.e., elements of $A$ or $c$). The goal of the OptiMUS system is to go from a natural language description of an optimization problem (e.g., a paragraph of text describing a business problem), which we call the \textit{problem description}, to a decision (assignment of numerical values to each variable) that solves the problem. 
\cledit{
We assume that the problem description contains all the information needed to model the optimization problem. In practice, users may give ambiguous or incomplete descriptions of an optimization problem (see \cite{wasserkrug2024large, lawless2024want} for a discussion). Helping users to resolve these ambiguities automatically 
is an exciting direction for future work.}

\begin{figure}
    \centering
    \includegraphics[width=0.6\textwidth]{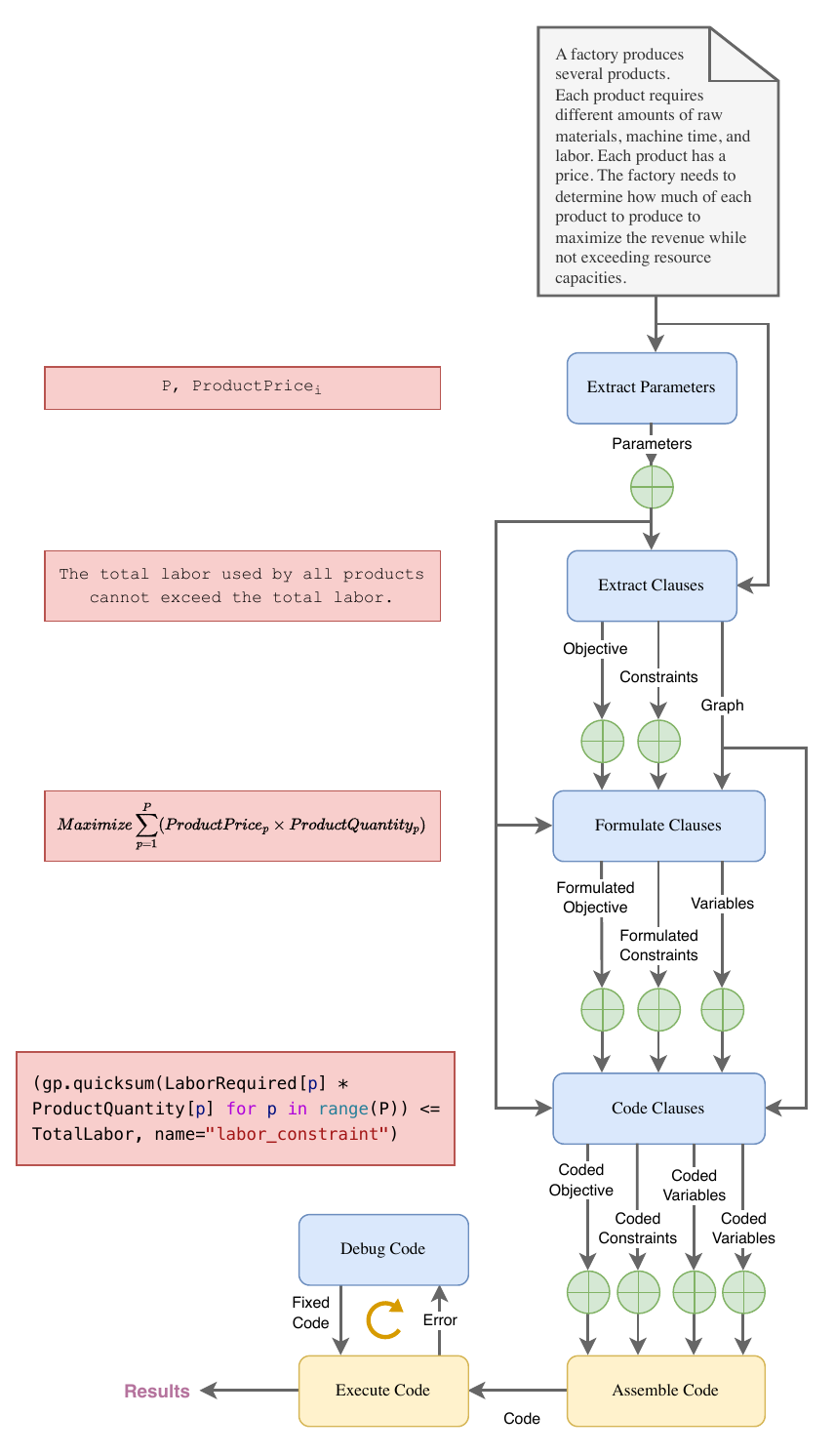}
    \caption{Optimus uses a sequential process with error correction (green circles) to model and solve optimization problems. It extracts the parameters and clauses, formulates them, generates the code for each, and finally synthesizes complete code and runs it. If the code raises an error, Optimus uses an iterative debugging loop to fix the code. 
    Red boxes provide sample outputs of each step of the pipeline. Blue boxes indicate LLM-based components, and yellow boxes indicate deterministic components.}
    \label{fig:flow}
\end{figure}

OptiMUS-0.3 decomposes these two tasks into a sequence of LLM-powered steps that incrementally constructs an optimization model and implements it. An overview of the complete workflow of OptiMUS-0.3 is outlined in \cref{fig:flow}. 
At a high level, OptiMUS-0.3 begins by 1) identifying parameters from the problem description
and writing a natural-language description of each clause in the problem. 
It then 2) formulates both the clauses and decision variables of the optimization model with mathematical precision, using \LaTeX~code. 
Finally, 3) OptiMUS-0.3 uses the \LaTeX~model of the optimization problem and the extracted parameters to generate code snippets for each clause in Python and assembles the snippets into a single runnable code file.
By running this file, OptiMUS-0.3 outputs an optimal solution and optimal objective value. 
Throughout this process, OptiMUS-0.3 maintains a \textit{state} that documents what is known about the optimization model, 
as well as a \textit{connection graph} that tracks which variables appear in which constraint (detailed in Section \ref{subsec:state_description}). The current state is supplied in the prompt of every LLM-powered component (detailed in Section \ref{subsec:llm_component}).
Each LLM-powered component of OptiMUS-0.3 also incorporates an Error Correction (EC) module (detailed in Section \ref{subsec:ec}) that catches common errors and improves the accuracy of the modeling and implementation.

Our current implementation writes code that uses Gurobi \citep{achterberg2019gurobi}, a leading MILP solver, and its associated Python API gurobipy. However, we have designed OptiMUS-0.3 to be easy to extend to other modeling languages and backends. We expect it to perform equally well with other solvers (e.g., CPLEX, SCIP) that have well-documented APIs, as well as with other modeling languages such as cvxpy \citep{diamond2016cvxpy}. 

Many of the advances in optimization solvers in the last several decades rely on the fact that solvers are faster when they can exploit particular structures within the optimization problem. One of the main tasks of an optimization expert is to identify these structures so as to choose an appropriate solver or parameter settings.
However, detecting useful structures in a given optimization problem can be
very challenging, since it requires both a deep understanding of the mathematical structure of the problem and the capabilities of existing solvers. In view of this challenge, OptiMUS-0.3 includes two specialized optimization modules: 1) a Structure Detection Agent, that detects variables (e.g., Indicator variables) and constraints (e.g., SOS constraints) with special structure that can be exploited by the solver, and 2) a customized optimization coding agent, that can leverage callbacks in the solver and perform basic column and constraint sifting to help improve the runtime of the code. These elements represent a first step towards automating more advanced decomposition algorithms with an automated modeling framework.

\cledit{OptiMUS-0.3 can be run as an end-to-end pipeline or through an interactive web application designed to facilitate human-system collaboration. The OptiMUS web application uses the same underlying components as the end-to-end framework but provides additional opportunities for users to audit and edit the outputs of each stage of the OptiMUS pipeline. We defer a deeper discussion of the system to Section \ref{webapp}, and mention specific interaction mechanisms available in the user interface earlier when relevant.}

The remainder of this section details the key components of the OptiMUS-0.3 framework. Section \ref{subsec:state_description} describes the underlying state and connection graph of the system. Section \ref{subsec:llm_component} describes a sample LLM component of the system. The remaining sections describe specialized modules that improve the performance of the system, including error correction and debugging (Section \ref{subsec:ec}), specialized modeling tools (Section \ref{subsec:structure_detection}), and advanced optimization coding (Section \ref{subsec:opt_coding}). 
OptiMUS-0.3, and all associated components, have been released as open-source code
and are publicly available to use through a web application, described in Section \ref{webapp}.

\subsection{State} \label{subsec:state_description}

OptiMUS-0.3 manages and modifies the solution using states saved in JSON format. The state consists of the following components:

\begin{itemize}
\item \textbf{Parameters}: OptiMUS-0.3 can choose a symbol for each parameter, infer its shape, and define the parameter if  it is not explicitly included in the problem statement. 
Numerical data from the problem statement is replaced by a symbol and stored separately for later use, 
ensuring that the parameter list remains concise and easy to include in future prompts.
\item \textbf{Clauses}: Each clause (objective or constraint) consists of a natural language description, a \LaTeX~formulation, and code implementing the clause.

\item \textbf{Variables}: Like parameters, each variable has a symbol, a shape, and a definition. 
Unlike parameters, each variable also has a type, which can be \textit{Continuous}, \textit{Integer}, or \textit{Binary}.


\item \textbf{Background}: The background is a short string that explains the real-world context for the problem. Including this string in every prompt helps improve common sense reasoning.

\item \textbf{Connection Graph}: The connection graph is a bipartite graph $G = (V,E)$ that links clauses to their associated parameters and variables. Formally, for every variable $x_j$ (clause $c_i$) in our formulation we create a node $v_{x_j}$ ($v_{c_i}$). In total, the connection graph consists of $n + m + 1$ nodes $v \in V$ corresponding to $n$ variables, $m$ constraints and the objective. We create an edge $e \in {E}$ between a variable node $v_{x_j}$ and a clause node $v_{c_i}$ if the variable participates in the clause (i.e., $(v_{x_j}, v_{c_i}) \in {E}$ if $a_{ij} \neq 0$). We build the connection graph iteratively by adding nodes and edges as more constraints and variables are formulated. This structure is utilized during the coding phase to ensure that only the relevant parameters and constraints are passed to the LLM.

\end{itemize}

\begin{figure}
    \centering
    \includegraphics[width=\textwidth]{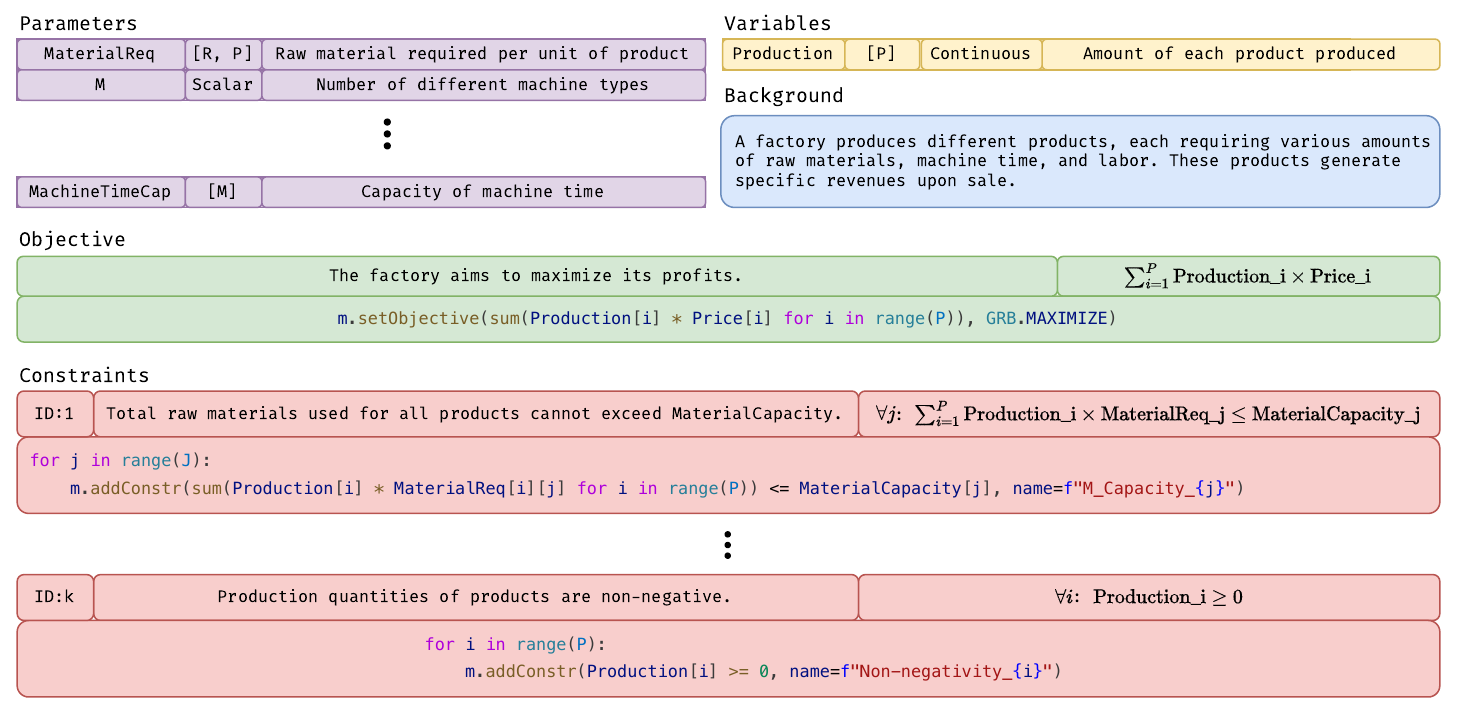}
    \caption{A completed state for a factory production optimization problem}
    \label{fig:state}
\end{figure}

The state is initialized as an empty object, and is completed step by step throughout the process. 
An example of the completed state for a factory optimization problem appears in \cref{fig:state}.

\subsection{Sample LLM Component} \label{subsec:llm_component}
OptiMUS-0.3 uses LLMs as a flexible tool to perform various tasks for the overall system, 
including extracting parameters, modeling clauses, and coding the components (see Figure \ref{fig:flow}). 
In the interest of brevity, this section outlines the design of a typical LLM component with the OptiMUS framework. 
Each LLM component is governed by a natural language directive called a \textit{prompt}. 
Within OptiMUS-0.3, each prompt includes the following key pieces of information:
\begin{itemize}
    \item \textbf{Task Description: } We describe the specific task, e.g., ``Your job is to extract natural language constraints for this paragraph defining an optimization problem''. 
    \item \textbf{Problem Context: } We include relevant information about the problem and the current solution in the prompt. During the formulation step, this involves the specific clause being addressed, along with all parameters and variables defined so far. During the coding step, this involves the clause and its formulation, supplemented by related parameters and variables which are dynamically loaded using the connection graph.
   
    \item \textbf{Examples: } We include a fixed set of sample outputs for the task in the prompt to help the LLM perform In-Context Learning \citep{dong2024surveyincontextlearning}. 
    For modules that output Python code, we also provide examples detailing the functionality of the gurobipy API.
\end{itemize}

For a comprehensive list of prompts used in OptiMUS-0.3 see our code\footnote{included in supplementary materials}. \cledit{Within the OptiMUS webapp, users can inspect and correct the output of every LLM-powered component of the system.}

\subsection{Error Correction} \label{subsec:ec}
To build a trustworthy and reliable system using LLMs, 
it is important to mitigate the impact of LLM hallucinations.
In the context of optimization modeling, 
an LLM might generate incorrect parameters, redundant mathematical constraints, 
or erroneous code. 
To address hallucinations, OptiMUS-0.3 uses two main error correction techniques: 
\textit{reflective prompts} and \textit{confidence-based user feedback}.

\subsubsection{Reflective prompts}
\begin{figure}
    \centering
\includegraphics{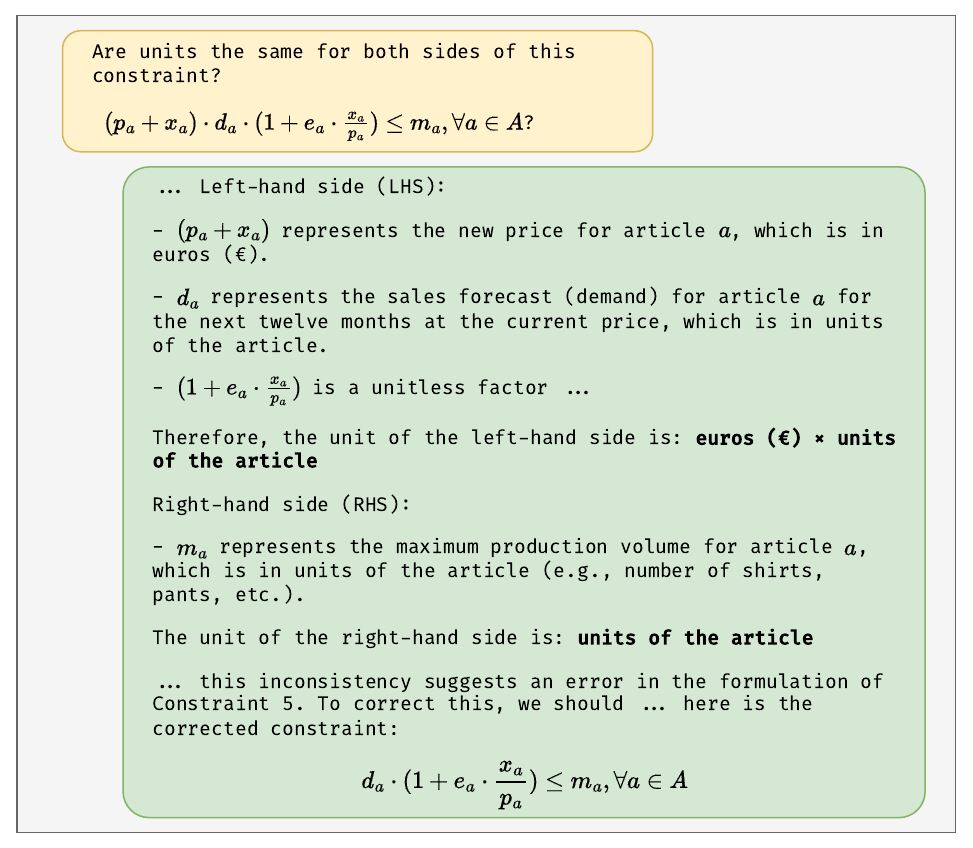}
    \caption{OptiMUS-0.3 can fix its constraint modeling errors when prompted ``\textit{Are units the same for both sides of $C$?}''}
    \label{fig:constraint-unit-check}
\end{figure}

LLMs can often identify and fix their mistakes by using reflective prompts \citep{shinn2023reflexion}, a strategy by which an LLM is asked to reflect upon its output and decide if it needs to be corrected. We analyzed the most common types of mistakes made by LLMs on the development dataset at each step of the process and designed reflective prompts to address each one (see Appendix \ref{app:reflex} for a full list of reflective prompts used in OptiMUS-0.3). 
Using these prompts significantly reduces the modeling error rate. For instance, Figure \ref{fig:constraint-unit-check} shows an example where the reflexive prompt \textit{Are units the same for both sides of this constraint?} can help correct a modeling error. The resulting improvement to OptiMUS is substantial (see \cref{table:ablation}), and substantiates the claim that 
domain-specific knowledge can be used to develop LLM applications that outperform general-purpose LLMs. 

\subsubsection{Confidence-based user/LLM feedback} \label{subsec:confidence}
In the context of real-world optimization modeling, it is exceedingly rare for a natural-language problem description to correspond to an unambiguous MILP formulation. Rather, the process of optimization modeling generally involves many cycles of formulation and user feedback to 
arrive at a correct model for the problem. OptiMUS-0.3 simulates this process in order to detect errors and improve its solve rate.
OptiMUS-0.3 can use a confidence-based reflective method to decide when to request help with its formulation, either from a user (if available) or another (more powerful) LLM.

	



\begin{figure}
    \centering
    \includegraphics[width=0.9\textwidth]{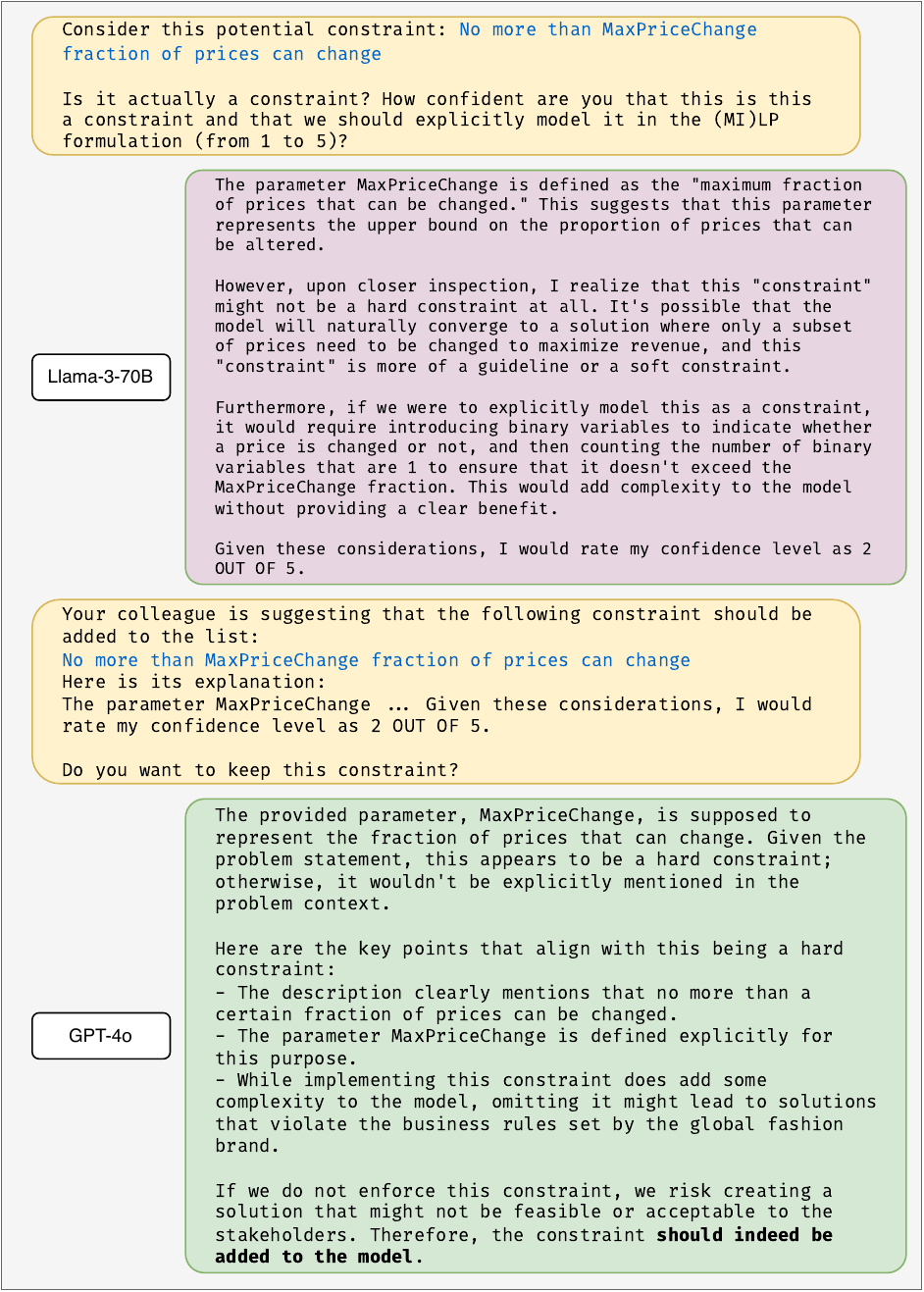}
    \caption{GPT-4o can provide feedback when Llama-3-70B lacks the confidence to identify a constraint.}
    \label{fig:llm_feedback}
\end{figure}

First, following the methodology of existing work on confidence calibration in LLMs \citep{virk2024enhancing},
we prompt the system to assess its confidence in its output on a scale of 1 to 5.
If the confidence score is less than 5, OptiMUS-0.3 asks the user (or a stronger LLM) to help.
The user can decide to either keep the output (e.g., constraint, code), remove it, or modify it.
\caledit{This confidence mechanism is applied across all three extraction and modeling modules: parameter extraction, variable identification, and constraint extraction.}
\cledit{In the OptiMUS webapp (\cref{webapp}), this feature is used to highlight outputs that merit particular attention.}
In our experiments (\cref{sec:experiments}), this feature is turned off by default to facilitate streamlined and reproducible testing without a user in the loop. 
See \cref{fig:llm_feedback} for an example of how a weaker LLM (Llama-3-70B) can call on a stronger LLM (GPT-4o) when it is not confident in its solution. 

\subsubsection{Debug Code} After the complete code is synthesized and executed by OptiMUS, any errors encountered during execution are passed to the LLM for debugging. At this stage, the LLM has the ability to modify any part of the code as needed. This step is particularly useful for resolving errors that arise from inconsistencies across individual code snippets, such as slight variations in variable names or formatting. The debugging process is repeated iteratively until the code executes successfully, with a maximum of five attempts.

\subsection{Structure Detection Agent} \label{subsec:structure_detection}

OptiMUS maintains a pool of optimization structures commonly used in optimization software. These structures can be specified explicitly in many modern optimization solver interfaces \citep{achterberg2019gurobi, manual1987ibmcplex, gamrath2016structure}, which can enhance problem-solving performance and simplify the code. Typical examples of structure include Special Ordered Set (SOS) {\citep{beale1976global}},  indicator variables, semi-continuous variables, and piecewise-linear constraints. At least one of these structures appears in approximately 10\% of the NLP4LP dataset (see Appendix \ref{app:structure_detection}). We expect these structures to be even more common, and thus important to exploit, in industrial applications.

\begin{figure}[t]
\centering
{\includegraphics[width=0.45\textwidth]{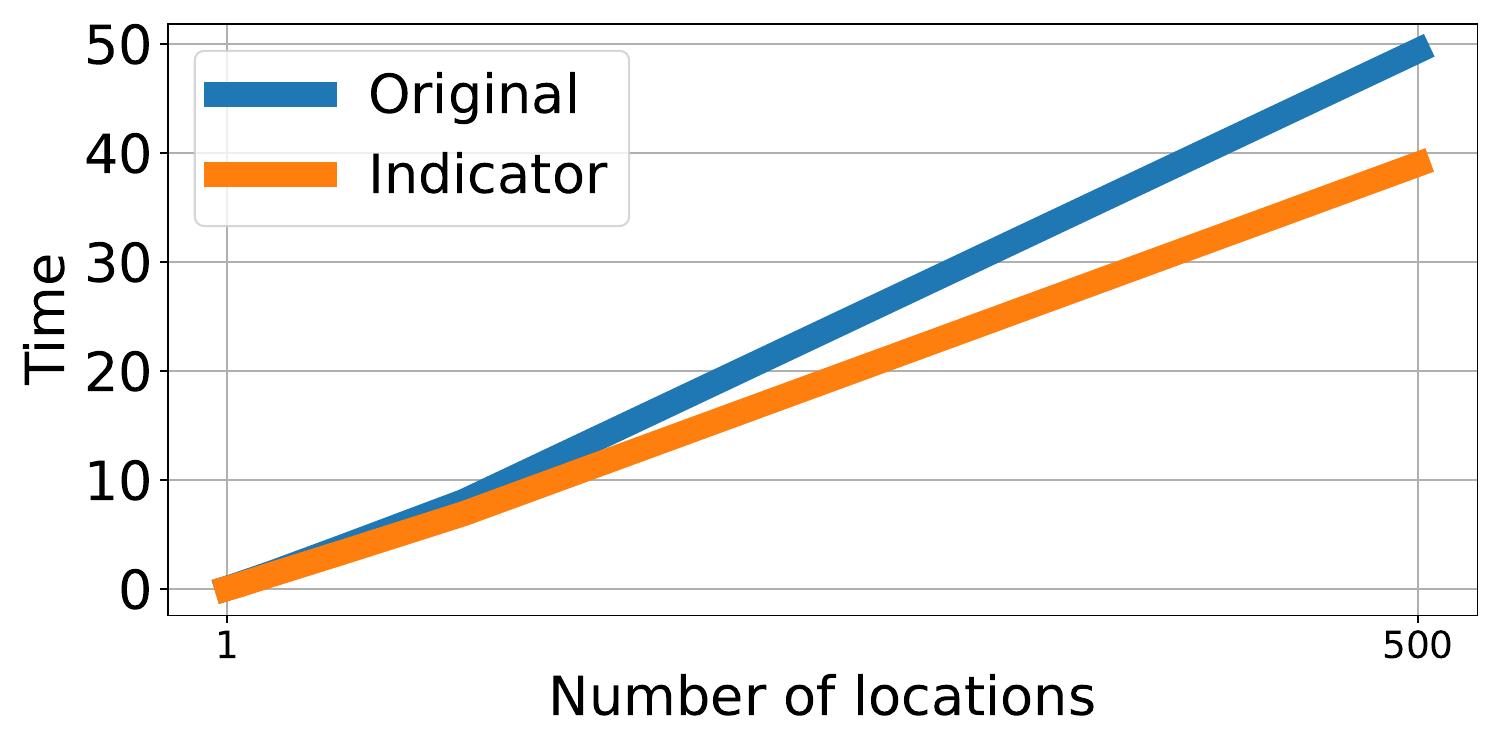}}
{\includegraphics[width=0.45\textwidth]{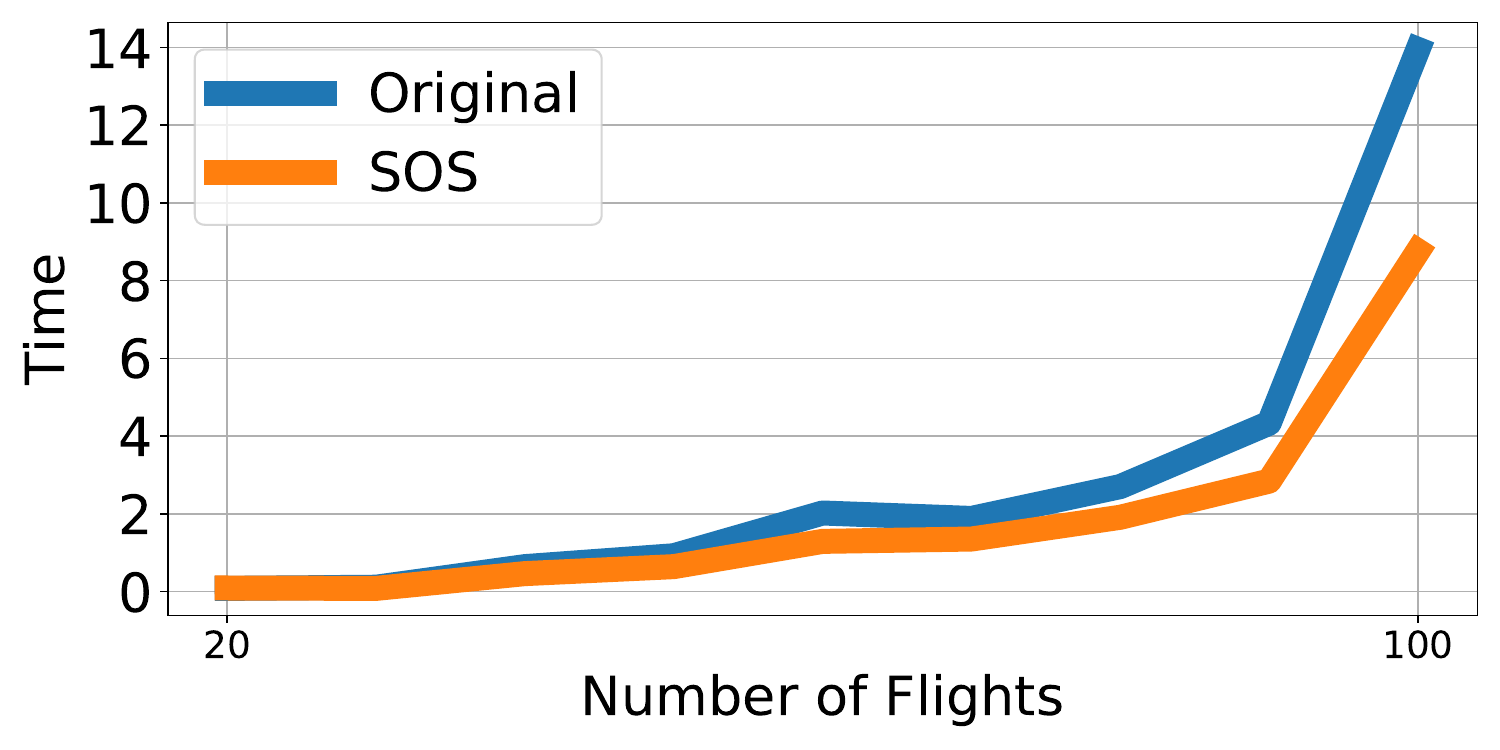}}
  \caption{Impact of Structure Detection on OptiMUS Performance. (Left) Speedup of solving a facility location optimization problem with indicator variables with the structure identified to the solver (indicator) versus with OptiMUS modeling of indicator variable. (Right) Speedup of solving a flight assignment problem with SOS-constraints with the structure identified to the solver (SOS) versus with OptiMUS modeling of the SOS constraints. Full details of problem instances are included in Appendix \ref{app:structure_detection}. \label{fig:structure_detect_performance}}
\end{figure}

To detect structure in a constraint (or variable), OptiMUS iterates through these structures and formats them into a structure detection prompt. Within each prompt, the LLM is
provided with the structure description, explained by an example illustrating how the structure should be exploited. The LLM is asked to decide
whether the structure is relevant to the existing formulation. Upon identifying the appropriate structure, the formulation is adjusted to highlight the problem structure. For instance, a set of constraints that indicate only one of a group of decision variables can be non-zero may be reformulated as type-1 SOS constraint. This structure is then conveyed to gurobi via the Python interfaces for special structures such as SOS constraints or indicator variables. This information can either be exploited within the branch and bound solver (e.g., using SOS constraints for branching rules) or allows the solver to automatically reformulate the structure into linear constraints (i.e., a big-M formulation for indicator constraints) \citep{gurobiReport}. While it may seem counter-intuitive to identify structure in a MILP only to have that structure re-formulated into linear constraints, in practice we found that automated methods within Gurobi can generate tighter formulations than OptiMUS alone (e.g., a better selection for big-M values). Figure \ref{fig:structure_detect_performance} shows two instances where identifying SOS constraints and indicator variables leads to faster performance than a naive implementation. Full details of these examples are included in Appendix \ref{app:structure_detection}. The structure detection agent is run during the formulate clauses stage of the algorithm.

OptiMUS also maintains a \textit{problem structure} pool that enumerates combinatorial optimization problems with special structure that can be exploited in fast custom solvers (e.g., SAT, network flow, routing problems). For instance, the traveling salesman problem (TSP) can be solved much more efficiently using a specialized solver such as Concorde as opposed to general-purpose MILP solvers \citep{cook2011traveling}. \cledit{OptiMUS iterates through this pool of special problem structures given an initial natural language description of the problem. When the optimization problem falls into one of the problem types in the structure pool, OptiMUS identifies the particular problem structure and suggests using a customized solver via a notification in the OptiMUS webapp.}

\subsection{Advanced Optimization Coding Agent} \label{subsec:opt_coding}
For extremely large-scale optimization problems, optimization solvers are often embedded into a high-level optimization framework and called as a subroutine: examples include decomposition algorithms such as column generation, Benders decomposition, and cutting plane methods. 
To leverage variable and constraint structure in this large-scale context, 
the subproblem solver must invoke an advanced solver interface (for example, call-backs, model attribute query and analysis). 
While we leave the problem of developing a fully automated system that can perform advanced decomposition algorithms to future work, we have added several features to OptiMUS that advance this vision.
OptiMUS implements a dedicated coding agent that exploits advanced solver functionality such as callbacks, and can iteratively call a solver within a simple variable or constraint sifting scheme (also known as delayed constraint generation). Neither of these modules requires explicitly reformulating the problem (e.g., generating a pricing problem within column generation), but allow for better computational performance than a naive implementation. 

Similar to the structure detection agent, the optimization coding agent maintains a series of template prompts for this functionality. Within each prompt, the LLM is provided with the purpose of the template. We illustrate this functionality with an application to sifting, a simple version of constraint or column generation, in \cref{sec:sifting}. The advanced coding agent runs during the Code Clauses stage of OptiMUS (see \cref{fig:flow}).

\subsubsection{Sifting} \label{sec:sifting}

Sifting is a large-scale linear optimization framework initially proposed in \citep{bixby1992very}. Consider the primal linear optimization problem
\[ \min_x \quad c^{\top} x \quad \text{subject to} \quad A x = b, \quad x \geq
   0, \]
where $A \in \mathbb{R}^{m \times n}$ and $n \gg m$. On the primal side,
sifting combines the idea of primal simplex and column generation by solving a
restricted master problem,
\[ \min_{x_S} \quad c_S^{\top} x_S \quad \text{subject to} \quad A_S x_S = b,
   \quad x_S \geq 0, \]
for a subset of columns $S \subseteq \{1,\ldots,n\}$. The restricted problem can be solved by any linear
programming subroutine. Its dual variable $y_S$ will be used to price out
dual infeasible columns
\[ I_S := \{ j : c_j - A_j^{\top} y < 0 \} . \]
Variable sifting works by iteratively updating $S \leftarrow S \cup I_S$
untill $S = \varnothing$. The idea can be generalized to the dual side when $m
\gg n$, where a restricted problem contains a subset of rows. We provide an example of constraint sifting in OptiMUS in \cref{fig:constraint-gen} on a security-constrained unit commitment problem. For details of the problem see Appendix \ref{app:scuc}. Most of the constraints in the instance are inactive, making it a good candidate for sifting.

\begin{figure}
\centering
	\includegraphics[scale=0.3]{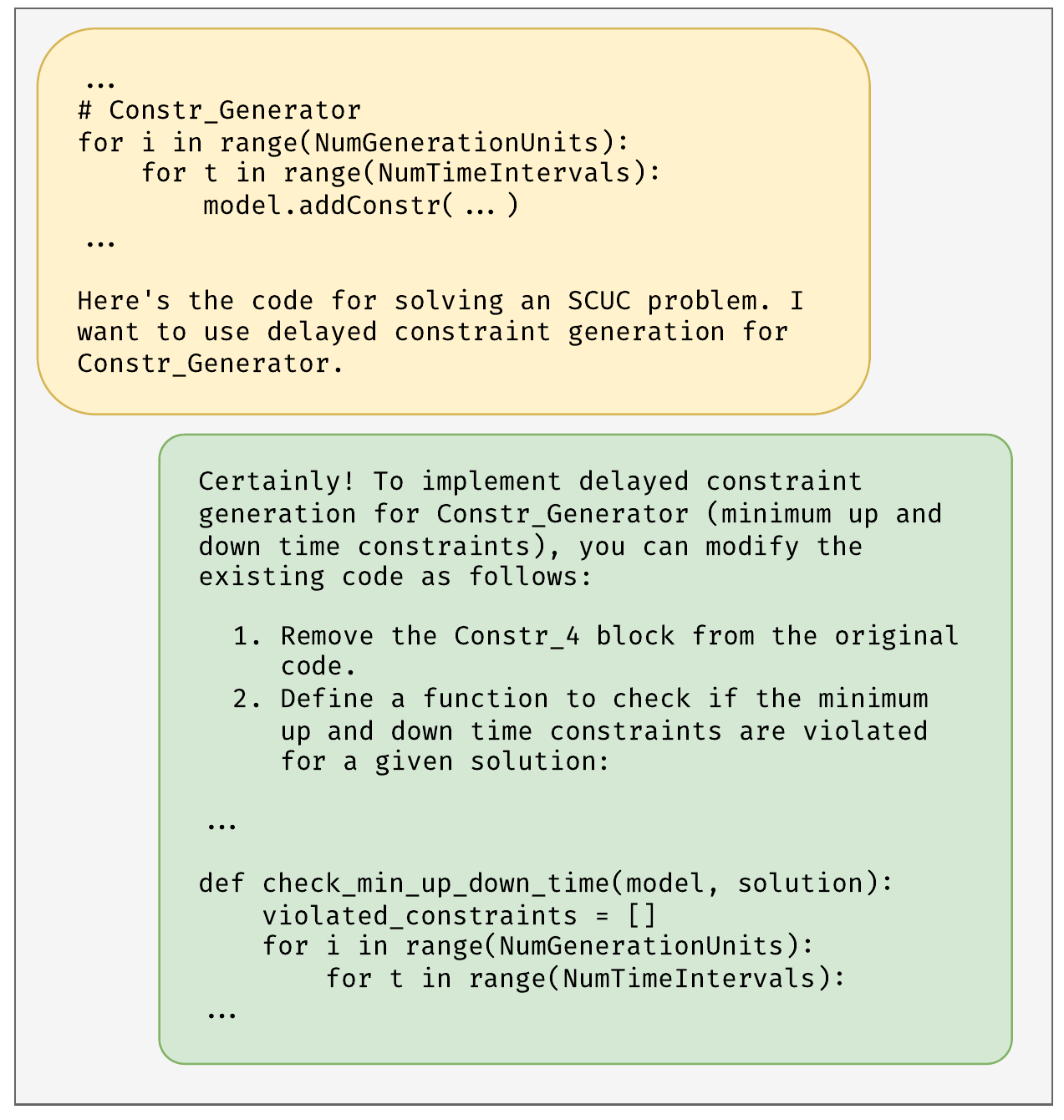}\includegraphics[scale=0.3]{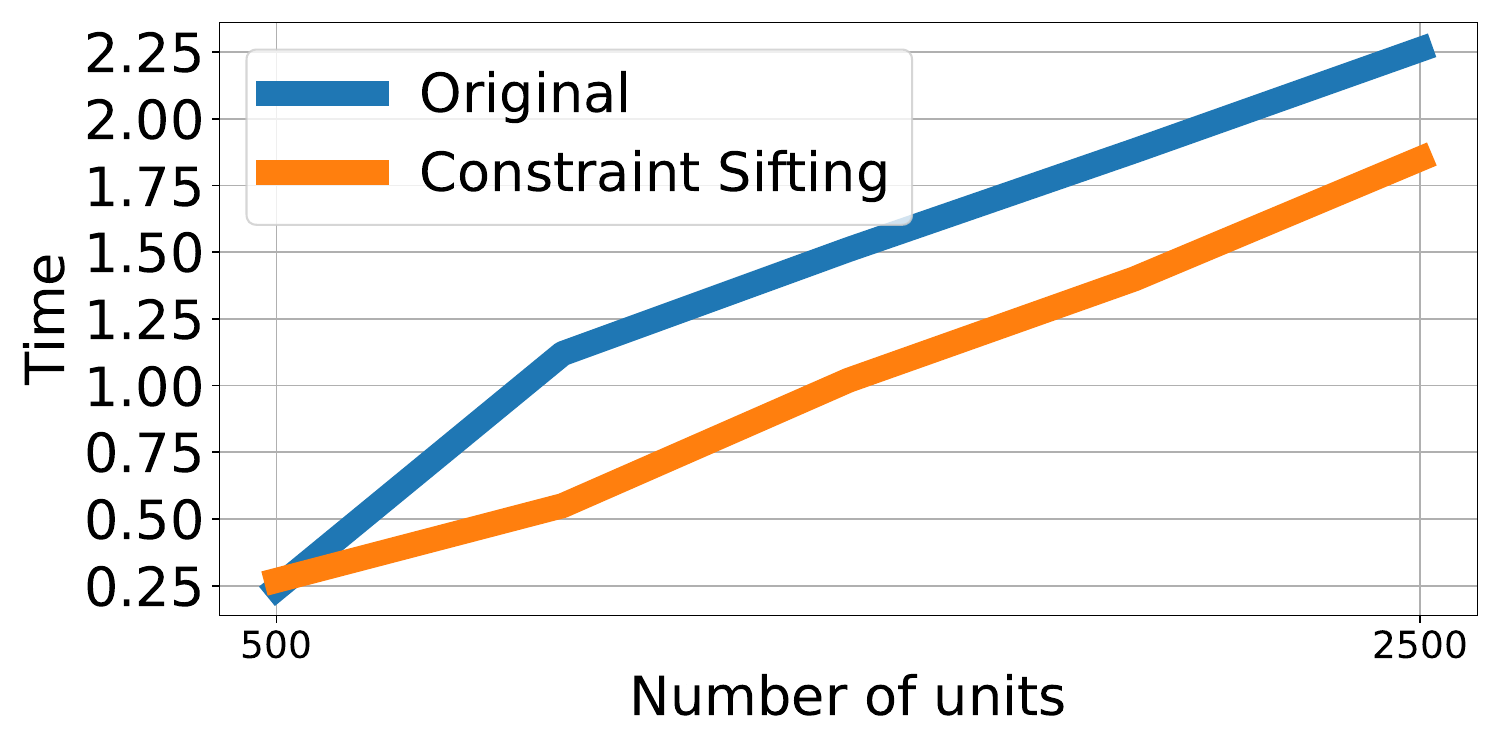}
\caption{Left: constraint sifting prompt. Right: performance plot for solving to a 5\% gap. \label{fig:constraint-gen}}
\end{figure}

\section{Experiments}
\label{sec:experiments}

In this section, we conduct a comprehensive evaluation of OptiMUS-0.3. We showcase the superior performance of OptiMUS-0.3 on \cledit{both the NLP4LP dataset and two existing benchmark datasets for optimization modeling} highlighting its strengths and weaknesses.
An ablation study demonstrates the impact of different system components on our results,
and a sensitivity analysis probes the internal dynamics of OptiMUS-0.3. 
We conclude this section by identifying failure cases and potential areas for further improvement.

\subsection{Overall Performance}

\paragraph{Baselines} To evaluate the overall performance of OptiMUS, we compare it with standard prompting, Reflexion, and Chain-of-Experts (CoE) \citep{shinn2023reflexion, xiao2023chain}. Reflexion is the highest-performing general-purpose framework, and \cledit{CoE is a recently proposed agentic framework for natural-language optimization modeling. We include self-reported results for two state-of-the-art auto-formulation methods based on fine-tuning LLMs specifically for optimization modeling: ORLM \citep{tang2024orlmtraininglargelanguage}, and LLMOPT \citep{jiang2024llmopt}. We also include results from the initial conference publication of this work, which we denote as OptiMUS-0.2. We report results for the OptiMUS-0.3 framework with both GPT-4o \citep{openai2023gpt4} and o3 \citep{jaech2024openai}, an LLM that automatically includes more advanced reasoning at the expense of longer inference times. Due to the cost of running o3, we report performance with standard prompting from \cite{song2026nemo} and only run the OptiMUS framework on datasets where o3 alone does not already achieve near-perfect accuracy. In our experiments, we use the default API parameters for all LLM calls; exact model versions and inference settings are reported in Appendix~\ref{app:inference_settings}.} 

\paragraph{Evaluation Metrics and Datasets} Three main metrics have been used to assess the accuracy of LP modeling tools in the literature: accuracy, compilation error (CE) rate, and runtime error (RE) rate. However, a method can generate an irrelevant short code that runs, or fix runtime and compilation errors by completely removing relevant sections of the code. Hence, we only compare the models' accuracy. Accuracy is defined as the number of instances correctly solved. An instance is considered as correctly solved only if 1) the code runs successfully, 2) the optimal value is correct (including infeasible instances), and 3) the optimal solution is correct.
\caledit{To check whether a generated solution matches the reference, we extract the numeric values of all decision variables from the solver output and compare them to the reference as multisets (unordered collections), making the check invariant to variable ordering. One limitation is that this requires the generated model to use the same variable space as the reference. A valid but differently-parameterized formulation (e.g., aggregate rather than per-unit variables) will produce a false negative, so condition~3 is a lower bound on true accuracy; condition~2 (optimal value match) serves as a softer complementary check.}
\caledit{Not all benchmark datasets include ground-truth optimal solutions. For datasets that provide only optimal values (NL4OPT and IndustryOR), we check for correctness by assessing execution accuracy, counting an instance as correct if the code runs successfully and the optimal value matches the reported value. For NLP4LP, which includes both optimal values and solutions, we apply the full three-condition accuracy defined above.} 

\cledit{In addition to the NLP4LP dataset introduced in this paper, we include results for two existing datasets for optimization modeling: NL4OPT \citep{ramamonjison2023nl4opt} and IndustryOR \citep{tang2024orlmtraininglargelanguage}. The latter dataset includes 100 real-world case studies from eight different industries.} \caledit{To isolate core modeling and solving capabilities, all methods in Table~\ref{table:performance-main} receive the same preprocessed input with parameter names identified. The OptiMUS system does not require preprocessed input: in fact, parameter extraction is the first pipeline step (Section~\ref{sec:methodology}), and the preprocessed input is equivalent to what OptiMUS's own extraction module produces. Rather, we provide preprocessed input to all methods to ensure a like-for-like comparison of modeling and solving ability.}

\begin{table}
\centering
\caption{Accuracy of auto-formulation methods on benchmark datasets. OptiMUS-0.2 requires an additional parameter conversion from the NLP4LP data files.
\label{table:performance-main}}
\small
\begin{tabular}{lccccccc}
\toprule
& & & \multicolumn{3}{c}{NLP4LP} & \\
\cmidrule(lr){4-6}
& LLM & NL4OPT & Easy & Hard & Case Studies & IndustryOR \\
\midrule
\rowcolor{gray!20} \multicolumn{7}{c}{\textit{Methods based on direct prompting}} \\
Standard   & GPT-4o & 47.3\%        & 34.6\%              & 20.3\%              & 14.3\%        & 28.0\% \\
Standard   & o3     & 97.3\%$^\dag$ & 62.7\%              & 42.4\%              & 28.6\%        & 44.0\% \\
Reflexion  & GPT-4o & 53.0\%        & 35.8\%              & 22.0\%              & 14.3\%        & --     \\
\midrule
\rowcolor{gray!20} \multicolumn{7}{c}{\textit{Methods based on fine-tuning LLMs}} \\
LLMOPT & Qwen1.5-14B  & 93.0\%$^*$ & \multicolumn{2}{c}{83.8\%$^{*\ddag}$} & --    & 46.0\%$^*$ \\
ORLM   & Deepseek-Math & 85.7\%$^*$ & 85.7\%             & 72.9\%              & 0.0\%         & 38.0\%$^*$ \\
\midrule
\rowcolor{gray!20} \multicolumn{7}{c}{\textit{Methods based on agentic frameworks}} \\
CoE         & GPT-4o & --     & 64.2\%       & 49.2\%       & --    & --     \\
OptiMUS-0.2 & GPT-4o & 78.8\% & 84.4\%       & 26.4\%       & --    & --     \\
OptiMUS-0.3 & GPT-4o & 86.6\% & 78.2\% & 38.5\% & 42.9\% & 37.0\% \\
OptiMUS-0.3 & o3     & --     & 79.9\% & 75.4\% & 28.6\% & 46.0\% \\
\bottomrule
\end{tabular}

{\smallskip\raggedright
\noindent $^*$~Self-reported from original paper; blanks indicate no results available.\\
\noindent $^\dag$~From \citet{song2026nemo}.\\
\noindent $^\ddag$~Reported on a combined sample of the Easy and Hard subsets; per-subset breakdown not available.\par}

\end{table}
\color{black}

\cledit{Results are presented in \cref{table:performance-main}. OptiMUS-0.3 outperforms all other direct prompting and agentic methods in all datasets by a large margin, beating the standard GPT-4o baseline by over 43\% on the easy NLP4LP subset and by over 18\% on the hard subset. This improvement persists even with more advanced reasoning models such as o3.
This impressive performance improvement highlights the importance of modularity and structure compared to a single prompt to solve complex problems using LLMs. }

\caledit{The OptiMUS framework also remains competitive with fine-tuning approaches. With o3, OptiMUS comes within 6\% of ORLM on the Easy subset and outperforms ORLM on the Hard, Case Studies, and IndustryOR benchmarks.
The modular design of an agentic system thus closes much of the performance gap between direct prompting and finetuning, giving practitioners a simpler alternative that requires no retraining when better models are released. The improved performance with o3 over GPT-4o underscores this benefit directly, as updating OptiMUS to use a newer model requires only a change to the API call, whereas finetuning-based methods must repeat the full training procedure. This generalization gap is most stark on the case-study instances, where ORLM achieves 72.9\% on the NLP4LP Hard subset but 0\% on the case studies, while OptiMUS-0.3 (GPT-4o) achieves 42.9\%, suggesting that fine-tuned models may not generalize to the complex, real-world MILP structure present in the case studies.} 

\begin{figure}
    \centering
    \includegraphics[width=0.75\linewidth]{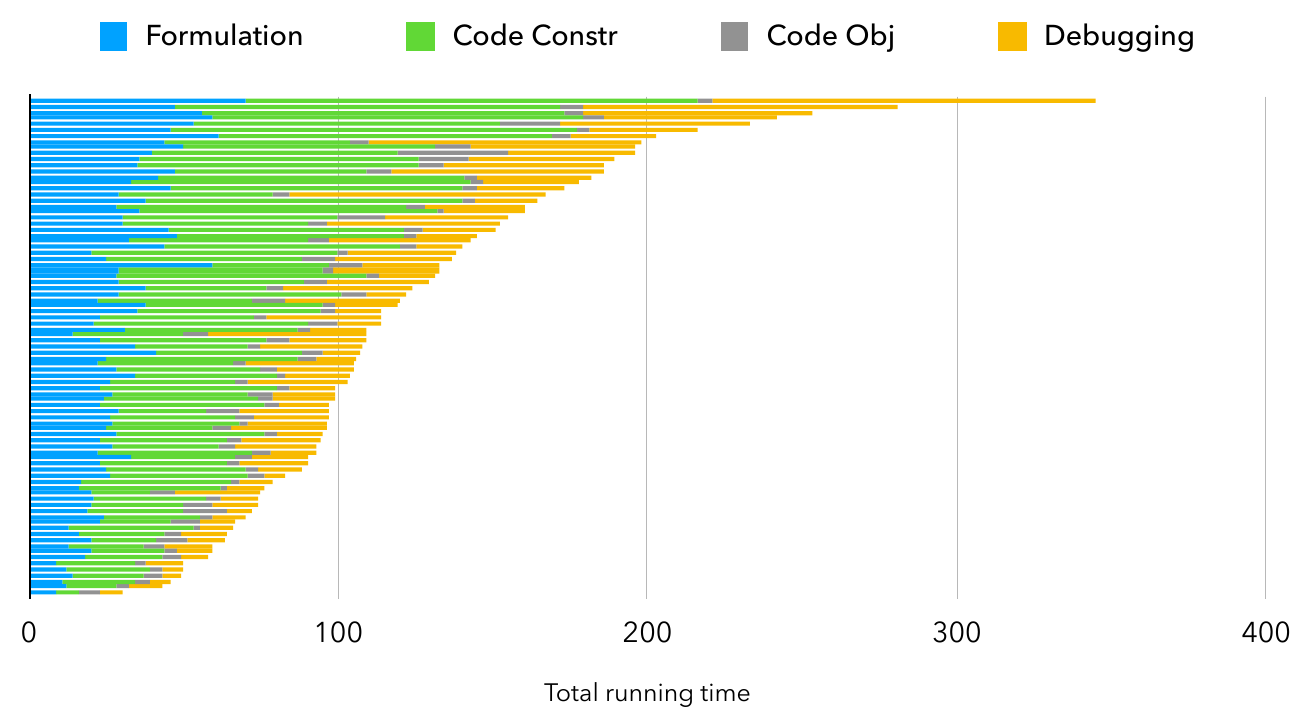}
    \caption{Distribution of running time (s) for different stages in the OptiMUS framework.}
    \label{fig:computation_time}
\end{figure}

\caledit{\paragraph{On Computation Time} In Figure \ref{fig:computation_time} we investigate the running time of the OptiMUS framework across 85 randomly selected instances of NLP4LP using GPT-4o and along different stages of the pipeline. The OptiMUS-0.3 framework takes no more than 350 seconds (median time 108 seconds) to model and solve any problem instance. For context, \citet{tang2024orlmtraininglargelanguage} report that human subject matter experts require on average approximately 150 \textit{minutes} to model and solve complex, real-world optimization problems of the kind collected in their IndustryOR benchmark; this figure should not be taken as representative of simpler, textbook-style instances. Coding and debugging represent the most time-consuming parts of the OptiMUS pipeline. Smaller LLM agents specialized to coding (e.g., codex) could further improve the computation time of the framework.
}


\caledit{\paragraph{On Stochasticity in LLM Outputs} LLM outputs are probabilistic, so calling the same model multiple times can produce different outputs. To evaluate the impact of this stochasticity on OptiMUS-0.3, we ran a stratified study: 15 randomly sampled easy instances and 15 hard instances, each run over 5 independent random seeds (150 total runs). This study is designed to measure \emph{consistency} across seeds, not to re-estimate overall accuracy; authoritative accuracy figures are in Table~\ref{table:performance-main}.}

\caledit{Table~\ref{table:stochasticity} summarizes the results. Easy instances are highly stable: 11 of 15 always pass across all five seeds, 4 of 15 show one or two seed-level failures, and none always fail. Hard instances show substantial stochasticity: 10 of 15 switch between passing and failing across seeds, giving a pass@5 rate (at least one pass in five seeds) of 73\%, well above the pass@1 rate of 35\% for this sample. Repeated sampling therefore does recover correct solutions for many hard instances. No instance was ever classified as ``incorrect'' (code runs but wrong answer), so all failures are identifiable execution errors. The error-correction modules strongly stabilize easy instances but only partially dampen stochasticity on harder problems, where small variations in intermediate steps can tip the pipeline toward failure.}

\begin{table}[ht]
\centering
\caption{Stochasticity robustness: 15 easy and 15 hard instances run over 5 seeds each. \emph{Always pass/fail}: consistent outcome across all 5 seeds. \emph{Mixed}: outcome varies across seeds. pass@5: fraction of instances passing on at least one of 5 seeds.
\label{table:stochasticity}}
\small
\begin{tabular}{lccccc}
\toprule
Stratum & Instances & pass@5 & Always pass & Mixed & Always fail \\
\midrule
Easy & 15 & 100\% & 11 & 4 & 0 \\
Hard & 15 &  73\% &  1 & 10 & 4 \\
\bottomrule
\end{tabular}
\end{table}

\caledit{\paragraph{On Calibration} A key feature of the OptiMUS-0.3 system is the confidence-based LLM feedback detailed in Section~\ref{subsec:confidence}. To evaluate whether these confidence scores are well-calibrated, we collected 410 annotated outputs across all three pipeline modules (parameter extraction, variable identification, and constraint extraction), covering both simple NLP4LP instances and the seven real-world case studies. A doctoral student in operations research manually annotated each extracted item as correct or incorrect. We report results in a type-I/type-II framing.}

\caledit{Type-I errors (high confidence, item wrong) occurred in 0 out of 402 cases where the LLM assigned confidence $\geq$4/5; the system never reported high confidence on an incorrect extraction across any module. Type-II errors (low confidence, item correct) occurred in 4 out of 8 low-confidence predictions (score $\leq$3/5), giving an over-flagging rate of 50\%. Among the 8 low-confidence items, accuracy was 25\% at score 1/5, 67\% at score 2/5, and 100\% at score 3/5, confirming that lower scores are genuinely predictive of errors. This asymmetry is intentional, as falsely trusting a wrong extraction (type-I) is far more costly for a practitioner-facing tool than unnecessarily reviewing a correct one (type-II). The 0\% type-I rate across 402 high-confidence predictions (spanning parameters, variables, and constraints on both simple and complex instances) provides strong evidence that the confidence mechanism is reliably conservative.}

\subsection{Ablation Study}

Table \ref{table:ablation} shows the impact of the choice of LLM on the performance of OptiMUS, broken down by its performance on the NLP4LP dataset. We evaluate the OptiMUS system with a leading closed-source LLM, GPT-4o, a reasoning model, o3, as well as a weaker open-source alternative, LLaMa-3.1-70B-Instruct \citep{meta2024llama3}. As expected, the performance of the OptiMUS-0.3 framework degrades when less capable LLMs, such as LLaMa-3.1-70B-Instruct, are used instead of GPT-4o or o3, especially on hard modeling tasks. Optimization tasks need complex reasoning, and smaller models may be better suited for simpler tasks.
However, the OptiMUS-0.3 framework with LLaMa-3.1-70B-Instruct outperforms GPT-4o with previous state-of-the-art prompting strategies, including Reflexion, on easy problems. 
We see that OptiMUS-0.3 is more robust to the quality of the underlying LLM, allowing weaker cheaper models to match the performance of stronger models on simple optimization modeling tasks.

\caledit{
Since the inception of the OptiMUS project, newer frontier models (e.g., OpenAI's GPT-5) have continued to improve code-and-reasoning capabilities, so the absolute accuracies in Table \ref{table:ablation} should be interpreted as a point-in-time snapshot rather than a fixed ceiling. Moreover, because NL4OPT and NLP4LP are public benchmarks, it becomes increasingly difficult to fully rule out partial benchmark exposure for later models, a known issue in LLM evaluation, so our main takeaway is deliberately model-agnostic: the modular structure of OptiMUS yields consistent gains using a strong general model (GPT-4o), a reasoning model (o3), and a weaker open model. 
We expect the same qualitative trend to hold for newer models, with diminishing improvements as benchmarks approach saturation.
}



Table \ref{table:ablation} also shows the impact of different components of the OptiMUS framework on its performance. \caledit{Each row removes one component from the full OptiMUS-0.3 pipeline while holding all others fixed, so the accuracy gap between that row and the full system quantifies the marginal contribution of that component.}

\caledit{Debugging (error correction on generated code) is the dominant driver of performance, reducing NLP4LP-Easy accuracy from 78.2\% to 26.7\% (--51.5pp) when removed. This effect is largest on hard instances, where LLMs frequently produce syntactically valid but semantically incorrect code; iterative solver feedback identifies and repairs these errors across multiple rounds. Extraction EC (error correction after clause extraction) accounts for the next largest contribution (--17.7pp, from 78.2\% to 60.5\%). It is especially important for hard instances, which involve multi-dimensional variables and complex indexing that the base extractor sometimes mishandles; crucially, missing constraints cannot be recovered later in the pipeline, so errors at this stage are final. Modeling EC (error correction after clause modeling) contributes a further --12.5pp (78.2\% to 65.7\%). Its effect is smaller than Extraction EC because many modeling errors can be recovered downstream in the debugging step. LLM Feedback contributes the smallest individual gain (--9.8pp, from 78.2\% to 68.4\%), catching residual cases where the solver returns a feasible but semantically incorrect solution. These effects are not independent (the full system's accuracy reflects their joint contribution), but the ablation rows make the marginal value of each component precise.}

\cledit{Table \ref{table:error_correction_confusion} further breaks down the impact of error correction methods on correct and incorrect items (i.e., constraints). In both constraint extraction and modeling, EC modules are able to fix a large fraction of errors without modifying most correct items.}
Note that we do not include the impact of the structure detection agent and advanced optimization coding agent as they do not impact the accuracy of the framework, but just the efficiency (i.e., runtime). For examples showcasing the impact of these components, see Appendix \ref{app:structure_detection}.


\cref{fig:ablation} (left) shows the relation between the performance and the number of debugging steps for OptiMUS with both GPT-4o and LLaMa-3.1-70B-Instruct. In both models, the first few debugging steps improve performance, but further steps do not result in improvement. 

\subsection{Failure Cases}
\cledit{We next analyze the most common reasons why OptiMUS fails. We categorized failure cases manually via a qualitative coding process based in grounded theory \citep{charmaz2006constructing}. We followed an open-coding process to analyze instances on which OptiMUS failed and tag the type of error. These initial codes were then grouped together into the following coarser categories: 
}

\begin{itemize}
    \item Extraction errors: OptiMUS extracts the wrong natural language constraints or objective (e.g., price is non-negative where price is a parameter), or fails to extract all of the constraints from the description.
    
    \item Formulation errors: OptiMUS formulates a mathematical model that does not match the natural language constraint (e.g., it uses the wrong parameter to lower-bound a variable).
    
    \item Coding errors: OptiMUS does not always generate error-free code even after debugging. Coding errors most often occur when the LLM is confused by the language used (e.g. a runtime error due to accessing indices that do not exist in a parameter array).
    
\end{itemize}

As shown in \cref{fig:ablation}, 
OptiMUS successfully extracts almost all the correct clauses for the easy dataset. This task is easy on the easy dataset because the easy instances are all LPs and involve only scalar values. 
However, for the hard instances --- comprising MILP problems and multi-dimensional variables --- 
parameter extraction and modeling are considerably more challenging than coding the resulting model. 
Note that the coding error rate is lower for the hard dataset because OptiMUS sometimes fails in modeling or formulating hard clauses, producing clauses that cannot be coded, which we do not include in the coding statistics.

\begin{table}
\centering
\caption{Ablation studies on OptiMUS-0.3 \label{table:ablation}}
\begin{tabular}{lcc}
\toprule
& NL4OPT & NLP4LP-Easy \\
\midrule
\multicolumn{1}{c}{\textbf{Importance of Different Components}} \\
\midrule
w/o Debugging 
& 73.2\% & 26.7\% \\

w/o Extraction EC  
& 86.7\% & 60.5\% \\ 

w/o Modeling EC  
& 83.8\% & 65.7\%  \\ 

w/o LLM Feedback 
& 86.6\% & 68.4\% \\

\textbf{OptiMUS-0.3} (GPT-4o) & \textbf{86.6\%} & \textbf{78.2\%} \\
\midrule
\multicolumn{1}{c}{\textbf{Performance with Different LLMs}} \\
\midrule
LLaMa3.1-70B-Instruct & 70.4\% & 31.5\% \\ 
GPT-4o & \textbf{86.6\%} & 78.2\% \\
\textbf{o3} & -- &\textbf{79.9\%} \\
\bottomrule
\end{tabular}
\end{table}

\begin{figure}[ht]
\centering
\begin{minipage}{0.49\textwidth}
\begin{tikzpicture}[scale=0.9]
\begin{axis}[
    xlabel={Number of Debugging Iterations},
    ylabel={Performance (\%)},
    xmin=1, xmax=6,
    ymin=0, ymax=100,
    xtick={1,2,3,4,5,6},
    ytick={0,20,40,60,80,100},
    legend pos=south east,
    ymajorgrids=true,
    grid style=dashed,
]

\addplot[
    color=main2,
    mark=o,
    style=dashed,
    ]
    coordinates {
    (1,73)(2,82)(3,87)(4,88)(5,88)(6,88)
    };
    \addlegendentry{GPT4-o Easy}
\addplot[
    color=main2,
    mark=o,
    ]
    coordinates {
    (1,26.7)(2,57.4)(3,71)(4,73)(5,73)(6,73)
    };
    \addlegendentry{GPT4-o Hard}

\addplot[
    color=main1,
    mark=square,
    style=dashed,
    ]
    coordinates {
    (1,2.2)(2,55.6)(3,77.3)(4,77.3)(5,77.3)(6,77.3)
    };
    \addlegendentry{LLaMa3-70B Easy}
\addplot[
    color=main1,
    mark=square,
    ]
    coordinates {
    (1,11.7)(2,35.2)(3,53.3)(4,53.3)(5,53.3)(6,53.3)
    };
    \addlegendentry{LLaMa3-70B Hard}
\end{axis}
\end{tikzpicture}
\end{minipage}
\begin{minipage}{0.49\textwidth}
\begin{tikzpicture}[scale=0.9]
\begin{axis}[
    ybar, 
    bar width=10pt,
    xlabel={Failure Type},
    legend style={at={(0.5,+0.95)},
      anchor=north,legend columns=-1},
    ylabel={Error rate (\%)}, 
    symbolic x coords={Extraction, Modeling, Coding},
    xtick=data,
    nodes near coords align={vertical},
    ymax=20,
    ymin=0,
    ytick={0,10,20,30,40,50,60,70,80,90,100}, 
    grid=major, 
    major grid style={dashed},
    xmajorgrids=false
    ]
\addplot+[ybar, fill=main1] plot coordinates {(Extraction,0.1) (Modeling, 3.8) (Coding,7.6)};
\addplot+[ybar, fill=main2] plot coordinates {(Extraction,15.8) (Modeling,12.6) (Coding,3.1)};

\legend{Easy,Hard}
\end{axis}
\end{tikzpicture}
\end{minipage}
\vspace{1em}
\caption{Left) Further debugging iterations improve performance. Right) For harder problems, most failures arise from clause extraction mistakes. For easier problems, most failures are due to coding errors. }
\label{fig:ablation}
\end{figure}

\begin{table}[t]
\caption{Error correction confusion matrices for constraint extraction (left) and modeling (right). Perfect performance is a diagonal matrix. \label{table:error_correction_confusion}}
\begin{center}
\begin{small}
\begin{minipage}{0.5\textwidth}
\centering
\begin{tabular}{lcc}
\toprule
& Not Modified & Modified \\
\midrule
Right & 219 & 7 \\
Wrong & 9 & 41 \\
\bottomrule
\end{tabular}
\end{minipage}%
\begin{minipage}{0.5\textwidth}
\centering
\begin{tabular}{lcc}
\toprule
& Not Modified & Modified \\
\midrule
Right & 231 & 2 \\
Wrong & 4 & 22 \\
\bottomrule
\end{tabular}
\end{minipage}
\end{small}
\end{center}
\end{table}

\subsection{\caledit{Scalability with Respect to Problem Complexity and Data Scale}} \label{sec:large-scale}

\caledit{We now provide fine-grainded analysis of the performance of OptiMUS on the seven case-study instances described in \cref{sec:dataset} (see \cref{table:case-studies}), which test two of the scalability dimensions introduced in \cref{intro}: \emph{problem complexity} (rich combinatorial structure, multiple classes of integer and binary variables, logical constraints) and \emph{data scale} (large-scale synthetic datasets generated to match the scales in each source paper). For each instance, we manually generate synthetic data and build and solve models according to the corresponding published paper.} \cref{table:model-comparison} reports the results. Although the version using o3 generates runnable code for five of the instances, only two of them are correct. On the other hand, GPT-4o generates runnable code for four of the instances, and three of them are correct. \caledit{These results demonstrate the challenge that problem complexity and data scale pose for current LLM-based modeling systems, and establish a concrete benchmark for future work on these dimensions.}

\begin{table}[ht]
\centering
\caption{Comparison of GPT-4o and o3 on real-world case studies.
Gap $= |{\rm Obj} - {\rm Ref. Obj.}| / (|{\rm Ref. Obj.}| + 1)$;
\textcolor{green!60!black}{\checkmark}~(0\% gap) means correct; \textcolor{red}{$\times$}~means runtime error or infeasible.}
\label{table:model-comparison}
\begin{tabular}{l r cc cc}
\toprule
& & \multicolumn{2}{c}{GPT-4o} & \multicolumn{2}{c}{o3} \\
\cmidrule(lr){3-4} \cmidrule(lr){5-6}
Instance ID & Ref. Obj. & Feasible & Gap & Feasible & Gap \\
\midrule
355 & 201.4  & \textcolor{green!60!black}{\checkmark} & \textcolor{green!60!black}{0\%} & \textcolor{green!60!black}{\checkmark} & \textcolor{green!60!black}{0\%} \\
356 & 851.7  & \textcolor{green!60!black}{\checkmark} & \textcolor{red}{97.5\%}  & \textcolor{green!60!black}{\checkmark} & \textcolor{red}{99.9\%}  \\
357 & 746.9  & \textcolor{red}{$\times$} & \textcolor{red}{$\times$} & \textcolor{red}{$\times$} & \textcolor{red}{$\times$} \\
358 & 1.0    & \textcolor{green!60!black}{\checkmark} & \textcolor{green!60!black}{0\%} & \textcolor{red}{$\times$} & \textcolor{red}{$\times$} \\
359 & 570.0  & \textcolor{red}{$\times$} & \textcolor{red}{$\times$} & \textcolor{green!60!black}{\checkmark} & \textcolor{red}{19.8\%} \\
360 & 820.0  & \textcolor{green!60!black}{\checkmark} & \textcolor{green!60!black}{0\%} & \textcolor{green!60!black}{\checkmark} & \textcolor{green!60!black}{0\%} \\
361 & 1217.4 & \textcolor{red}{$\times$} & \textcolor{red}{$\times$} & \textcolor{green!60!black}{\checkmark} & \textcolor{red}{99.7\%} \\
\midrule
\multicolumn{2}{l}{\textit{Correct}}   & \multicolumn{2}{c}{3 / 7} & \multicolumn{2}{c}{2 / 7} \\
\multicolumn{2}{l}{\textit{Feasible}}  & \multicolumn{2}{c}{4 / 7} & \multicolumn{2}{c}{5 / 7} \\
\bottomrule
\end{tabular}
\end{table}

\paragraph{Detailed analysis.} To better understand the failed instances, we inspected each incorrect or infeasible instance. The errors fall into three broad categories. \textit{Runtime errors} arise from implementation mistakes that prevent execution entirely: for instance~359, GPT-4o passes arc data stored as lists directly to Gurobi as variable keys, causing a type error since Gurobi requires hashable (tuple) indices; for instance~361, GPT-4o iterates over period indices starting from zero while the data keys begin at the actual starting period, producing a key error. \textit{Wrong formulations} occur when the model runs but misrepresents the problem structure: for instance~356, GPT-4o incorrectly defines excess as a binary per-student variable rather than a continuous per-(group, class) variable, and computes a quadratic deviation term incompatible with MIP solvers; for instance~361, o3 collapses the entire multi-period pool-management LP into a single scalar release decision per period, discarding the pool-state dynamics and multiorder-hit structure that drive the objective. \textit{Incomplete or placeholder models} represent cases where the LLM fails to engage with the problem at all: for instance~356, o3 ignores all class and capacity data and minimizes a dummy group-index objective; for instance~357, GPT-4o adds contradictory balance constraints that render the model infeasible, while o3 completely ignores the optimization objective; for instance~358, o3 applies the DEA normalization constraint over all decision-making units instead of the evaluated unit, and leaves weight variables unbounded below, making the problem unbounded; for instance~359, o3 omits mandatory detour constraints for long block sides, traffic restrictions, and forbidden paths, finding a shorter but infeasible route. \\

Together, these failures highlight three key challenges for LLMs on real-world OR problems: correctly handling domain-specific data structures, faithfully translating complex multi-indexed formulations into code, and recognizing when a problem requires rich auxiliary structure beyond a simple objective and basic constraints.

\caledit{\paragraph{Data scale.} A key design feature of OptiMUS is that problem data is stored externally and referenced symbolically: the LLM sees only a structured description of parameter names and shapes, not the raw data values. As a result, the prompt length is independent of input data size. To validate this design, we compare OptiMUS against a \emph{naive prompting baseline} that provides the full problem data as a JSON blob in the prompt alongside the description.} \caledit{We use the two case-study instances whose GPT-4o solutions are correct (Amazon Locker, instance 355; R-VRP, instance 360) and vary the input data size across different scales (i.e., by increasing the planning horizon for the locker problem and number of routes for VRP). For each (instance, scale) pair we generate 5 independent data instances (different random seeds), run both methods, and report \emph{correct-formulation rate}: the fraction of seeds where the generated code runs and finds a high-quality objective. We use a 600-second solver budget for all runs and define success as matching the reference objective within 1\%.}

\caledit{For instance 355 (Amazon Locker), OptiMUS maintains 100\% correct-formulation rate at all 8 scale levels (data 3\,KB--935\,KB): the same generated code is applied to data of any size. Naive prompting degrades from 80\% at $T{=}5$ to 40\% at $T{=}30$, indicating genuine formulation errors as description and data diverge, and exceeds GPT-4o's 128K-token context window at $T{\geq}75$ (data $\geq$245\,KB), making it infeasible beyond that point.} \caledit{For instance 360 (R-VRP), OptiMUS achieves 100\% correct-formulation rate at all 9 scale levels: it generates code once and applies it to data ranging from 2\,KB to 385\,KB. Naive prompting matches this at small to medium scales (data 2\,KB--190\,KB), achieving 80\% for two instances (one formulation error per seed out of five) and 100\% at all other tested scales. At large scales (272--385\,KB), naive prompting cannot generate code at all because the prompt exceeds GPT-4o's context window; OptiMUS continues to run unaffected.}

\caledit{Together, these results show that naive prompting has two distinct failure modes (formulation errors and context overflow at large scales), neither of which affects OptiMUS. The results further demonstrate that OptiMUS's data-separation architecture allows it to remain operational at data scales that are simply unreachable for naive prompting.}

\begin{figure}[ht]
    \centering
    \includegraphics[width=0.92\textwidth]{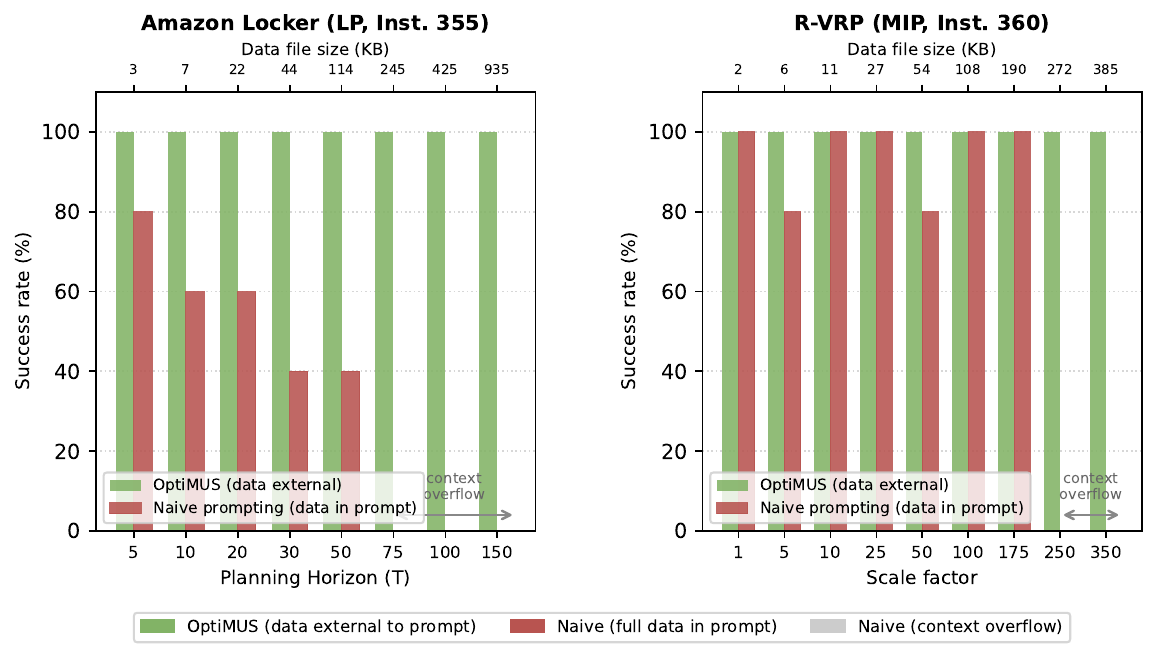}
    \caption{\caledit{Correct-formulation rate (5 seeds per scale) for OptiMUS vs.\ naive prompting on instance~355 (LP, left) and instance~360 (MIP, right). Hatched bars indicate context overflow; OptiMUS, which stores data externally, is unaffected at all scales.}}
    \label{fig:data_scale}
\end{figure}




\section{Web Application}
\label{webapp}
\begin{figure}
    \centering
    \includegraphics[width=0.8\textwidth]{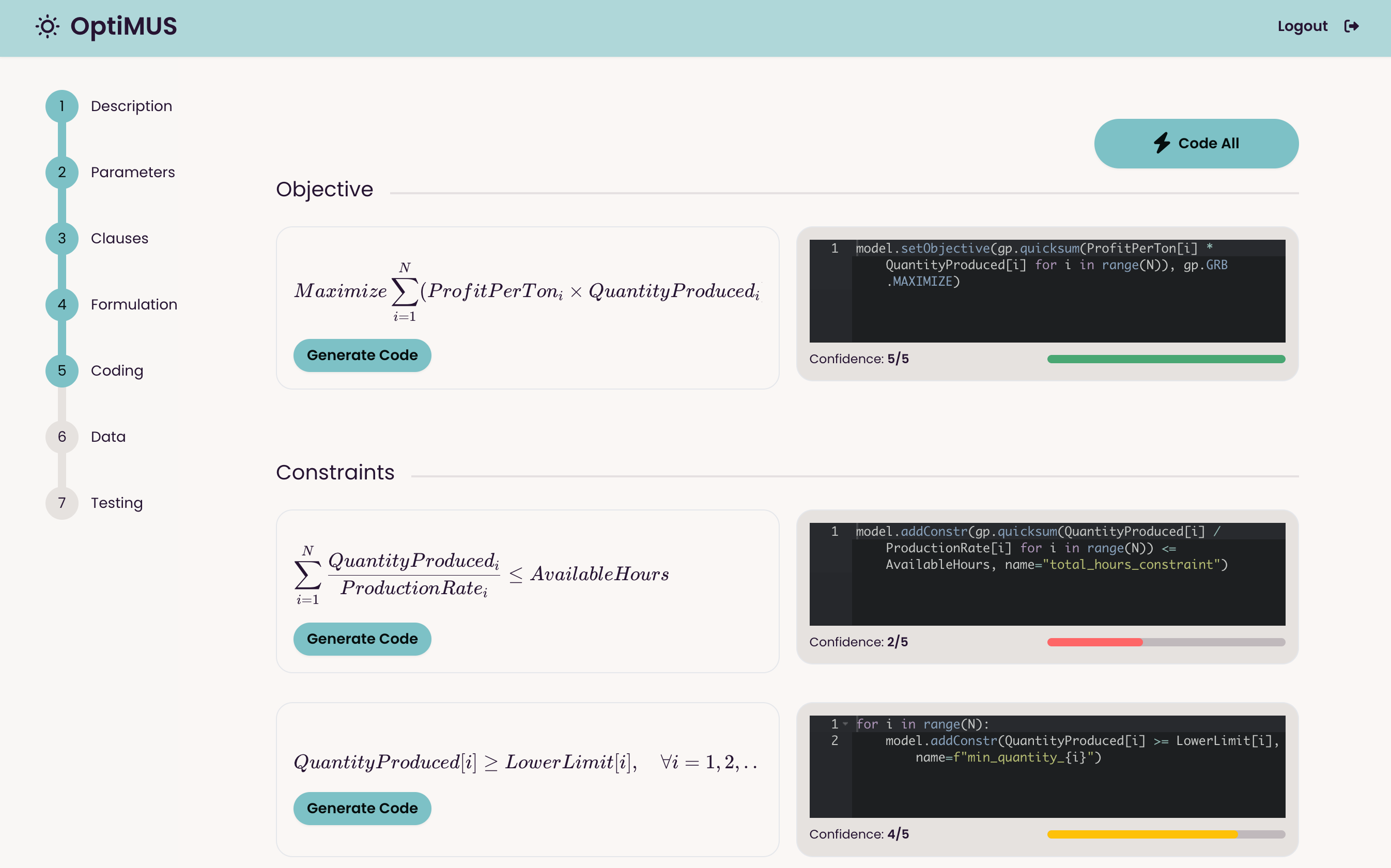}
    \caption{Sample view of the OptiMUS webapp during the coding phase in which individual clauses are translated to Python code. Users have the ability to see the confidence scores, correct the code directly in the panes to the right, or ask OptiMUS to regenerate code.}
    \label{fig:webapp_screenshot}
    \end{figure}

As discussed in \ref{intro}, the OptiMUS project aims to help optimization practitioners develop and maintain their models with ease, 
reducing development time and the risk of errors.
To further these goals, we designed and built a user-centric web app with an intuitive interface that enables users to interact seamlessly with the system and leverage the advantages of large language models (LLMs). The webapp allows users to follow the overall process of modeling an optimization model including extracting parameters and clauses, modeling each clause, and finally coding it up. Importantly, users can correct any output of the framework, including natural language clauses, \LaTeX~model, and code, to provide better supervision over the entire process. By allowing users to observe and provide input throughout each step, the web app brings significant speed and convenience to the modeling process while reducing the risk of errors. Figure \ref{fig:webapp_screenshot} shows a sample model in the Coding stage. Further details of all the different stages and features incorporated in the webapp can be found in Appendix \ref{app:webapp}. The webapp is publicly available at \href{https://optimus-solver.com/}{https://optimus-solver.com/}.

\section{Conclusion}
\label{conclusion}

How can we leverage LLMs to achieve complex goals? This paper interrogates this question in the domain of optimization
and showcases the importance of modular structure:
our ablation studies show that
debugging alone accounts for nearly half of performance gains on hard problems, 
and modular architecture allows weaker models to match stronger models with naive prompting.
We instantiate these principles in OptiMUS-0.3, a modular LLM-based agent that achieves state-of-the-art performance on NLP4LP, a new comprehensive dataset of challenging optimization problems, and scales to problems with long descriptions and complex real-world data.
\caledit{The primary lesson of this work is a design principle: in LLM-based optimization modeling, system architecture can enable weaker models to match stronger ones and agentic approaches to compete with fine-tuned specialist models without retraining costs.}

\caledit{OptiMUS is designed as a productivity tool for optimization practitioners: it accelerates the modeling and implementation workflow for users who already understand the problem domain, in the same way that GitHub Copilot accelerates software development for engineers who already know how to code.}
Real-world optimization problems are often complex and multifaceted.
Developing LLM-based solutions for these problems requires domain-specific considerations, including integrating existing optimization techniques to leverage problem structure. We are at the early stages of this research, but anticipate significant developments that will enable these systems to address more complex, industrial-level problems. It is interesting to notice that the challenge of using AI for an applied domain is much larger in safety-critical domains such as self-driving, which demand extremely high accuracy, than in domains where AI can function as an assistant and where answers can be checked by humans, as in theorem-proving or optimization. Here, AI systems with moderate accuracy can still usefully augment human effort.


\paragraph{\textbf{Future directions.}}
\caledit{The OptiMUS system presented here serves as a prototype for more powerful AI-assisted optimization modeling, and a step towards making classical optimization tools accessible to a broader range of users. Much more work remains to make such a system reliable enough to reduce the expertise required to deploy optimization in practice.}
While some improvements can be expected as language models grow more powerful,
we expect that substantial breakthroughs by the optimization community will be required to address 
many of the following challenges: 
\begin{itemize}[leftmargin=10pt]
    \item \textbf{Reliability:} Optimization algorithms generally offer guaranteed reliability. How can a system like OptiMUS reliably solve large-scale problems using unreliable LLMs that suffer hallucinations, overconfidence, and instability? 
    What kinds of guarantees can be made, and how expensive might these be?
    How should developers of such a system assess whether the system has succeeded in producing a correct model or not?
    \item \textbf{Trust:} A system like OptiMUS should help domain experts understand the model that was selected and decide whether it meets their needs. How should the system communicate a model back to the human user to facilitate effective feedback about the deficiencies in the model, or trust in a model that is correct? How should the system integrate this feedback? How can the system developer assess the effectiveness of this feedback and improve the system to enable quicker and easier feedback cycles?
    \item \textbf{Ambiguity:} Most optimization problems specified in natural language do not correspond unambiguously to a single optimization model. Rather, the process of formalizing the objective and constraints reduces the vagueness so that it is possible to distinguish a solution from any other proposal. A system like OptiMUS should help decision-makers think through their goals (which may be vague) to choose a model formulation by clarifying implicit requirements, eliciting the relative importance of different objectives, and searching internal or external databases for relevant data.
    \item \textbf{Fast Solvers:} Rapid decisions require a fast solver: possibly a well-chosen solver for the problem, or possibly even a custom solver written for the problem. A system like OptiMUS should be able to choose between available heuristics and parameters for provably optimal solvers using all information available about the problem to determine a solution method that meets runtime and accuracy requirements, or even to write a custom solver when it is warranted. To choose or write the best solver, we expect that the system will likely make use of natural-language information about the problem setting (using an LLM) as well as structured information about problem parameters (using a graph neural network). 
    \item \textbf{Larger datasets.} All of these research directions depend on the availability of appropriate datasets to develop, hone, and test new ideas. Most of these directions require a dataset of optimization problems consisting of a natural-language description coupled with problem data and solution code. While there are many excellent datasets for developing optimization algorithms, these are generally not associated with natural-language descriptions; and problems associated with natural language descriptions are generally either small (e.g., textbook problems) or lack associated problem data and solutions (e.g., published papers on applications of MIPs) as these elements are often proprietary. Semi-synthetic approaches to generating larger datasets are promising \citep{tang2024orlmtraininglargelanguage}, but creating MIP problems that are large-scale, realistic, and feasible is a challenge. Curating a larger dataset of optimization problems could be a key enabler driving further research and development in this area.
    \item \textbf{Beyond MILPs.} OptiMUS currently uses gurobipy as a modeling language. We made this choice as the wealth of online examples allows LLMs to successfully model many problems in gurobipy. However, many optimization problems are best expressed in a different framework: for example, Concorde for traveling salesman problems \citep{cook2011traveling}, cvxpy for control problems \citep{diamond2016cvxpy}, or minizinc for constraint programming problems \citep{nethercote2007minizinc}. Integrating these alternative modeling languages and identifying, for a given problem, which modeling language to use (or whether the problem is likely to be successfully modeled in the OptiMUS framework at all) is an important future direction.
    
\end{itemize}


    
    







\section*{Acknowledgments}
AA, WG, CL, and MU gratefully acknowledge support from
the National Science Foundation (NSF) Award IIS-2233762, 
the Office of Naval Research (ONR) Awards N000142212825, 
N000142412306, 
and N000142312203, 
the Alfred P. Sloan Foundation,
and from IBM Research as a founding member of Stanford Institute for Human-centered Artificial Intelligence (HAI).
HB thanks the Aker Scholarship Foundation for financial support. ST gratefully acknowledges support from the Air Force Office of Scientific Research (AFOSR) under Grant FA9550-23-1-0251 and in part by the Office of Naval Research under Grant N00014-24-1-2164.
This manuscript was improved by feedback from anonymous referees.
\bibliographystyle{plainnat}
\bibliography{references}

\begin{APPENDIX}{}
\section{Detailed Solution Statistics}

\newcommand{\cmark}{\ding{51}}
\newcommand{\xmark}{\ding{55}}

\begin{table}[h]
\centering
\caption{GPT-4o per-instance results on NLP4LP (instances 1--354). $\dagger$\,=\,Hard (instances 290--354). \cmark\,=\,correct, \xmark\,=\,failed, infeas\,=\,infeasible/unbounded.}
\label{tab:per_instance}
\scriptsize
\setlength{\tabcolsep}{3pt}
\begin{tabular}{cc|cc|cc|cc|cc|cc|cc|cc}
\toprule
ID & Status & ID & Status & ID & Status & ID & Status & ID & Status & ID & Status & ID & Status & ID & Status \\
\midrule
1 & \cmark & 46 & \cmark & 91 & \cmark & 136 & \cmark & 181 & infeas & 226 & \cmark & 271 & \cmark & $316^{\dagger}$ & \xmark \\
2 & \cmark & 47 & \cmark & 92 & \cmark & 137 & \cmark & 182 & \cmark & 227 & \xmark & 272 & \cmark & $317^{\dagger}$ & infeas \\
3 & infeas & 48 & \xmark & 93 & \cmark & 138 & \cmark & 183 & \cmark & 228 & \cmark & 273 & \cmark & $318^{\dagger}$ & \xmark \\
4 & \xmark & 49 & \cmark & 94 & \cmark & 139 & \xmark & 184 & infeas & 229 & \cmark & 274 & \cmark & $319^{\dagger}$ & \cmark \\
5 & \xmark & 50 & \cmark & 95 & \cmark & 140 & \xmark & 185 & \xmark & 230 & \xmark & 275 & \cmark & $320^{\dagger}$ & \cmark \\
6 & \xmark & 51 & infeas & 96 & \cmark & 141 & \xmark & 186 & \xmark & 231 & \xmark & 276 & infeas & $321^{\dagger}$ & \cmark \\
7 & \xmark & 52 & \cmark & 97 & \xmark & 142 & \cmark & 187 & \cmark & 232 & \xmark & 277 & infeas & $322^{\dagger}$ & \xmark \\
8 & \cmark & 53 & \xmark & 98 & \xmark & 143 & \cmark & 188 & \xmark & 233 & \cmark & 278 & \cmark & $323^{\dagger}$ & \cmark \\
9 & \cmark & 54 & \cmark & 99 & \cmark & 144 & \cmark & 189 & \cmark & 234 & \cmark & 279 & \cmark & $324^{\dagger}$ & \cmark \\
10 & \cmark & 55 & \cmark & 100 & \cmark & 145 & \xmark & 190 & \cmark & 235 & \cmark & 280 & \cmark & $325^{\dagger}$ & \cmark \\
11 & \cmark & 56 & \cmark & 101 & \cmark & 146 & \cmark & 191 & \cmark & 236 & \cmark & 281 & \cmark & $326^{\dagger}$ & infeas \\
12 & \cmark & 57 & infeas & 102 & \cmark & 147 & \cmark & 192 & \cmark & 237 & infeas & 282 & \cmark & $327^{\dagger}$ & \cmark \\
13 & \cmark & 58 & infeas & 103 & infeas & 148 & \cmark & 193 & \cmark & 238 & infeas & 283 & \cmark & $328^{\dagger}$ & \cmark \\
14 & \xmark & 59 & \cmark & 104 & \cmark & 149 & \cmark & 194 & \cmark & 239 & \cmark & 284 & \cmark & $329^{\dagger}$ & \cmark \\
15 & \cmark & 60 & \cmark & 105 & \cmark & 150 & \xmark & 195 & \cmark & 240 & infeas & 285 & \cmark & $330^{\dagger}$ & \xmark \\
16 & \cmark & 61 & \xmark & 106 & \cmark & 151 & \xmark & 196 & \cmark & 241 & \cmark & 286 & \cmark & $331^{\dagger}$ & \xmark \\
17 & \cmark & 62 & \cmark & 107 & \cmark & 152 & \xmark & 197 & \cmark & 242 & \cmark & 287 & \cmark & $332^{\dagger}$ & \xmark \\
18 & \cmark & 63 & \cmark & 108 & \cmark & 153 & \cmark & 198 & \cmark & 243 & \xmark & 288 & \cmark & $333^{\dagger}$ & \xmark \\
19 & \cmark & 64 & \cmark & 109 & \cmark & 154 & \cmark & 199 & \cmark & 244 & infeas & 289 & \cmark & $334^{\dagger}$ & \xmark \\
20 & \cmark & 65 & \cmark & 110 & \cmark & 155 & \cmark & 200 & \cmark & 245 & \xmark & $290^{\dagger}$ & \cmark & $335^{\dagger}$ & \xmark \\
21 & \cmark & 66 & \cmark & 111 & \cmark & 156 & \xmark & 201 & \cmark & 246 & \cmark & $291^{\dagger}$ & \cmark & $336^{\dagger}$ & \xmark \\
22 & \cmark & 67 & \cmark & 112 & \cmark & 157 & \xmark & 202 & \cmark & 247 & \cmark & $292^{\dagger}$ & \cmark & $337^{\dagger}$ & \xmark \\
23 & \cmark & 68 & infeas & 113 & \cmark & 158 & \xmark & 203 & \cmark & 248 & \cmark & $293^{\dagger}$ & \xmark & $338^{\dagger}$ & \xmark \\
24 & \cmark & 69 & \xmark & 114 & \cmark & 159 & \cmark & 204 & \cmark & 249 & \cmark & $294^{\dagger}$ & \cmark & $339^{\dagger}$ & \xmark \\
25 & \cmark & 70 & \cmark & 115 & \cmark & 160 & \xmark & 205 & \cmark & 250 & \xmark & $295^{\dagger}$ & \cmark & $340^{\dagger}$ & \xmark \\
26 & \cmark & 71 & \xmark & 116 & \cmark & 161 & \cmark & 206 & \cmark & 251 & infeas & $296^{\dagger}$ & \cmark & $341^{\dagger}$ & \xmark \\
27 & \cmark & 72 & \xmark & 117 & \cmark & 162 & \xmark & 207 & \cmark & 252 & \cmark & $297^{\dagger}$ & infeas & $342^{\dagger}$ & \xmark \\
28 & infeas & 73 & \xmark & 118 & \xmark & 163 & \cmark & 208 & \cmark & 253 & \xmark & $298^{\dagger}$ & \cmark & $343^{\dagger}$ & \xmark \\
29 & \cmark & 74 & \cmark & 119 & \cmark & 164 & \cmark & 209 & \cmark & 254 & \cmark & $299^{\dagger}$ & \xmark & $344^{\dagger}$ & \xmark \\
30 & \cmark & 75 & \cmark & 120 & \xmark & 165 & \cmark & 210 & \cmark & 255 & \cmark & $300^{\dagger}$ & \cmark & $345^{\dagger}$ & \xmark \\
31 & \cmark & 76 & \cmark & 121 & \cmark & 166 & \cmark & 211 & \cmark & 256 & \cmark & $301^{\dagger}$ & \xmark & $346^{\dagger}$ & \xmark \\
32 & \cmark & 77 & infeas & 122 & \cmark & 167 & \cmark & 212 & \cmark & 257 & \xmark & $302^{\dagger}$ & \cmark & $347^{\dagger}$ & \xmark \\
33 & \cmark & 78 & \xmark & 123 & infeas & 168 & \xmark & 213 & \xmark & 258 & \cmark & $303^{\dagger}$ & \xmark & $348^{\dagger}$ & \xmark \\
34 & \cmark & 79 & infeas & 124 & \cmark & 169 & \cmark & 214 & \xmark & 259 & \cmark & $304^{\dagger}$ & \xmark & $349^{\dagger}$ & \xmark \\
35 & \cmark & 80 & \cmark & 125 & \cmark & 170 & \xmark & 215 & \cmark & 260 & \xmark & $305^{\dagger}$ & \xmark & $350^{\dagger}$ & \xmark \\
36 & \cmark & 81 & \cmark & 126 & infeas & 171 & infeas & 216 & \xmark & 261 & infeas & $306^{\dagger}$ & \xmark & $351^{\dagger}$ & \xmark \\
37 & \cmark & 82 & \cmark & 127 & \cmark & 172 & \xmark & 217 & \xmark & 262 & \xmark & $307^{\dagger}$ & \cmark & $352^{\dagger}$ & \xmark \\
38 & \cmark & 83 & infeas & 128 & \cmark & 173 & \cmark & 218 & \cmark & 263 & \cmark & $308^{\dagger}$ & \xmark & $353^{\dagger}$ & \xmark \\
39 & \cmark & 84 & \cmark & 129 & \cmark & 174 & \cmark & 219 & infeas & 264 & \cmark & $309^{\dagger}$ & \xmark & $354^{\dagger}$ & \xmark \\
40 & \cmark & 85 & \cmark & 130 & \cmark & 175 & \cmark & 220 & \cmark & 265 & \cmark & $310^{\dagger}$ & \xmark &   &   \\
41 & \cmark & 86 & \cmark & 131 & \cmark & 176 & \xmark & 221 & \xmark & 266 & infeas & $311^{\dagger}$ & infeas &   &   \\
42 & \xmark & 87 & \xmark & 132 & \xmark & 177 & \xmark & 222 & \cmark & 267 & infeas & $312^{\dagger}$ & infeas &   &   \\
43 & \cmark & 88 & \xmark & 133 & \cmark & 178 & \xmark & 223 & \cmark & 268 & \xmark & $313^{\dagger}$ & \xmark &   &   \\
44 & \xmark & 89 & \cmark & 134 & \cmark & 179 & infeas & 224 & infeas & 269 & \cmark & $314^{\dagger}$ & \xmark &   &   \\
45 & \cmark & 90 & \xmark & 135 & infeas & 180 & \cmark & 225 & \xmark & 270 & \xmark & $315^{\dagger}$ & infeas &   &   \\
\bottomrule
\end{tabular}
\end{table}

\begin{table}[h]
\centering
\caption{O3 per-instance results on NLP4LP (instances 1--354). $\dagger$\,=\,Hard (instances 290--354). \cmark\,=\,correct, \xmark\,=\,failed, infeas\,=\,infeasible/unbounded.}
\label{tab:per_instance_o3}
\scriptsize
\setlength{\tabcolsep}{3pt}
\begin{tabular}{rc|rc|rc|rc|rc|rc|rc|rc}
\toprule
ID &Status& ID &Status& ID &Status& ID &Status& ID &Status& ID &Status& ID &Status& ID &Status\\
\midrule
1 & \cmark & 46 & \cmark & 91 & \cmark & 136 & \cmark & 181 & infeas & 226 & \cmark & 271 & \cmark & $316^{\dagger}$ & \xmark \\
2 & \cmark & 47 & \cmark & 92 & \cmark & 137 & \cmark & 182 & \cmark & 227 & \xmark & 272 & \cmark & $317^{\dagger}$ & infeas \\
3 & infeas & 48 & \xmark & 93 & \cmark & 138 & \cmark & 183 & \cmark & 228 & \cmark & 273 & \cmark & $318^{\dagger}$ & \xmark \\
4 & \xmark & 49 & \cmark & 94 & \cmark & 139 & \xmark & 184 & infeas & 229 & \cmark & 274 & \cmark & $319^{\dagger}$ & \cmark \\
5 & \cmark & 50 & \cmark & 95 & \cmark & 140 & \xmark & 185 & \xmark & 230 & \xmark & 275 & \cmark & $320^{\dagger}$ & \xmark \\
6 & \xmark & 51 & infeas & 96 & \xmark & 141 & \xmark & 186 & \xmark & 231 & \xmark & 276 & infeas & $321^{\dagger}$ & \cmark \\
7 & \xmark & 52 & \cmark & 97 & \xmark & 142 & \cmark & 187 & \cmark & 232 & \xmark & 277 & infeas & $322^{\dagger}$ & \cmark \\
8 & \cmark & 53 & \xmark & 98 & \xmark & 143 & \cmark & 188 & \cmark & 233 & \cmark & 278 & \cmark & $323^{\dagger}$ & \cmark \\
9 & \cmark & 54 & \cmark & 99 & \cmark & 144 & \cmark & 189 & \cmark & 234 & \cmark & 279 & \cmark & $324^{\dagger}$ & \cmark \\
10 & \cmark & 55 & \cmark & 100 & \cmark & 145 & \xmark & 190 & \cmark & 235 & \cmark & 280 & \cmark & $325^{\dagger}$ & \cmark \\
11 & \cmark & 56 & \cmark & 101 & \cmark & 146 & \cmark & 191 & \cmark & 236 & \cmark & 281 & \cmark & $326^{\dagger}$ & infeas \\
12 & \cmark & 57 & infeas & 102 & \cmark & 147 & \cmark & 192 & \cmark & 237 & infeas & 282 & \cmark & $327^{\dagger}$ & \cmark \\
13 & \cmark & 58 & infeas & 103 & infeas & 148 & \cmark & 193 & \cmark & 238 & infeas & 283 & \cmark & $328^{\dagger}$ & \cmark \\
14 & \xmark & 59 & \cmark & 104 & \cmark & 149 & \cmark & 194 & \cmark & 239 & \cmark & 284 & \cmark & $329^{\dagger}$ & \cmark \\
15 & \cmark & 60 & \cmark & 105 & \cmark & 150 & \xmark & 195 & \cmark & 240 & infeas & 285 & \cmark & $330^{\dagger}$ & \cmark \\
16 & \cmark & 61 & \xmark & 106 & \cmark & 151 & \xmark & 196 & \cmark & 241 & \cmark & 286 & \cmark & $331^{\dagger}$ & \xmark \\
17 & \cmark & 62 & \cmark & 107 & \cmark & 152 & \xmark & 197 & \cmark & 242 & \cmark & 287 & \cmark & $332^{\dagger}$ & \cmark \\
18 & \cmark & 63 & \cmark & 108 & \cmark & 153 & \cmark & 198 & \cmark & 243 & \xmark & 288 & \cmark & $333^{\dagger}$ & \cmark \\
19 & \cmark & 64 & \cmark & 109 & \cmark & 154 & \cmark & 199 & \xmark & 244 & infeas & 289 & \cmark & $334^{\dagger}$ & \cmark \\
20 & \cmark & 65 & \cmark & 110 & \cmark & 155 & \cmark & 200 & \cmark & 245 & \xmark & $290^{\dagger}$ & \cmark & $335^{\dagger}$ & \cmark \\
21 & \cmark & 66 & \cmark & 111 & \cmark & 156 & \xmark & 201 & \cmark & 246 & \cmark & $291^{\dagger}$ & \cmark & $336^{\dagger}$ & \cmark \\
22 & \cmark & 67 & \cmark & 112 & \cmark & 157 & \xmark & 202 & \cmark & 247 & \cmark & $292^{\dagger}$ & \cmark & $337^{\dagger}$ & \cmark \\
23 & \cmark & 68 & infeas & 113 & \cmark & 158 & \cmark & 203 & \cmark & 248 & \cmark & $293^{\dagger}$ & \cmark & $338^{\dagger}$ & \xmark \\
24 & \cmark & 69 & \xmark & 114 & \cmark & 159 & \cmark & 204 & \cmark & 249 & \cmark & $294^{\dagger}$ & \cmark & $339^{\dagger}$ & \cmark \\
25 & \cmark & 70 & \cmark & 115 & \cmark & 160 & \xmark & 205 & \cmark & 250 & \cmark & $295^{\dagger}$ & \cmark & $340^{\dagger}$ & \cmark \\
26 & \cmark & 71 & \xmark & 116 & \cmark & 161 & \cmark & 206 & \cmark & 251 & infeas & $296^{\dagger}$ & \cmark & $341^{\dagger}$ & \cmark \\
27 & \cmark & 72 & \xmark & 117 & \cmark & 162 & \xmark & 207 & \cmark & 252 & \cmark & $297^{\dagger}$ & infeas & $342^{\dagger}$ & \cmark \\
28 & infeas & 73 & \xmark & 118 & \xmark & 163 & \cmark & 208 & \cmark & 253 & \xmark & $298^{\dagger}$ & \cmark & $343^{\dagger}$ & \cmark \\
29 & \cmark & 74 & \cmark & 119 & \cmark & 164 & \cmark & 209 & \cmark & 254 & \cmark & $299^{\dagger}$ & \xmark & $344^{\dagger}$ & \cmark \\
30 & \cmark & 75 & \cmark & 120 & \xmark & 165 & \cmark & 210 & \cmark & 255 & \cmark & $300^{\dagger}$ & \cmark & $345^{\dagger}$ & \xmark \\
31 & \cmark & 76 & \cmark & 121 & \cmark & 166 & \cmark & 211 & \cmark & 256 & \cmark & $301^{\dagger}$ & \cmark & $346^{\dagger}$ & \xmark \\
32 & \cmark & 77 & infeas & 122 & \cmark & 167 & \cmark & 212 & \cmark & 257 & \cmark & $302^{\dagger}$ & \cmark & $347^{\dagger}$ & \cmark \\
33 & \cmark & 78 & \xmark & 123 & infeas & 168 & \xmark & 213 & \xmark & 258 & \cmark & $303^{\dagger}$ & \xmark & $348^{\dagger}$ & \cmark \\
34 & \cmark & 79 & infeas & 124 & \cmark & 169 & \cmark & 214 & \cmark & 259 & \cmark & $304^{\dagger}$ & \xmark & $349^{\dagger}$ & \xmark \\
35 & \cmark & 80 & \cmark & 125 & \cmark & 170 & \xmark & 215 & \cmark & 260 & \xmark & $305^{\dagger}$ & \xmark & $350^{\dagger}$ & \cmark \\
36 & \cmark & 81 & \cmark & 126 & infeas & 171 & infeas & 216 & \xmark & 261 & infeas & $306^{\dagger}$ & \cmark & $351^{\dagger}$ & \cmark \\
37 & \cmark & 82 & \cmark & 127 & \cmark & 172 & \xmark & 217 & \xmark & 262 & \xmark & $307^{\dagger}$ & \cmark & $352^{\dagger}$ & \cmark \\
38 & \cmark & 83 & infeas & 128 & \cmark & 173 & \cmark & 218 & \cmark & 263 & \cmark & $308^{\dagger}$ & \xmark & $353^{\dagger}$ & \cmark \\
39 & \cmark & 84 & \cmark & 129 & \cmark & 174 & \cmark & 219 & infeas & 264 & \cmark & $309^{\dagger}$ & \cmark & $354^{\dagger}$ & \cmark \\
40 & \cmark & 85 & \cmark & 130 & \cmark & 175 & \cmark & 220 & \cmark & 265 & \cmark & $310^{\dagger}$ & \xmark &   &   \\
41 & \cmark & 86 & \cmark & 131 & \cmark & 176 & \xmark & 221 & \xmark & 266 & infeas & $311^{\dagger}$ & infeas &   &   \\
42 & \xmark & 87 & \xmark & 132 & \xmark & 177 & \xmark & 222 & \cmark & 267 & infeas & $312^{\dagger}$ & infeas &   &   \\
43 & \cmark & 88 & \xmark & 133 & \cmark & 178 & \xmark & 223 & \cmark & 268 & \xmark & $313^{\dagger}$ & \xmark &   &   \\
44 & \xmark & 89 & \cmark & 134 & \cmark & 179 & infeas & 224 & infeas & 269 & \cmark & $314^{\dagger}$ & \xmark &   &   \\
45 & \cmark & 90 & \xmark & 135 & infeas & 180 & \cmark & 225 & \xmark & 270 & \cmark & $315^{\dagger}$ & infeas &   &   \\
\bottomrule
\end{tabular}
\end{table}

\newpage

\section{Terms and Definitions}\label{app:terms_and_definitinos}

\newcolumntype{L}[1]{>{\raggedright\arraybackslash}p{#1}}

\begin{longtable}{@{}L{3cm} L{12cm}@{}}
\caption{Definitions of Terms Used in OptiMUS-0.3}
\label{tab:definitions} \\

\toprule
\multicolumn{2}{@{}l}{\textbf{Term: Clause}} \\
\midrule
Definition: & A constraint or objective within the optimization problem. \\
\cmidrule(l){1-2}
How We Use Them: & We refer to constraints and objectives as clauses to simplify terminology. \\
\cmidrule(l){1-2}
Example: & The constraint $x + y \leq 10$ is a clause. \\
\addlinespace
\midrule[0.75pt]\midrule[0.75pt]

\multicolumn{2}{@{}l}{\textbf{Term: Connection graph}} \\
\midrule
Definition: & A graph that records which variables and parameters appear in each constraint. \\
\cmidrule(l){1-2}
How We Use Them: & Used to ensure consistency of formulations and focus the LLM on relevant context. \\
\cmidrule(l){1-2}
Example: & Constraint $C1$ connects to variables $x$ and $y$ in the connection graph. \\
\addlinespace
\midrule[0.75pt]\midrule[0.75pt]

\multicolumn{2}{@{}l}{\textbf{Term: State}} \\
\midrule
Definition: & The collection of all parameters, clauses, variables, and background information managed and modified during the solution process. \\
\cmidrule(l){1-2}
How We Use Them: & Saved and updated in JSON format throughout the problem-solving steps. \\
\cmidrule(l){1-2}
Example: & The state includes parameters like demand, variables like production quantity, and clauses like capacity constraints. \\
\addlinespace
\midrule[0.75pt]\midrule[0.75pt]

\multicolumn{2}{@{}l}{\textbf{Term: Reflective prompts}} \\
\midrule
Definition: & Prompts designed to encourage the LLM to reflect on and correct its own mistakes. \\
\cmidrule(l){1-2}
How We Use Them: & Used to reduce modeling errors by having the LLM check and fix its outputs. \\
\cmidrule(l){1-2}
Example: & Asking "Are units the same for both sides of constraint $C$?" \\
\addlinespace
\midrule[0.75pt]\midrule[0.75pt]

\multicolumn{2}{@{}l}{\textbf{Term: Confidence-based user feedback}} \\
\midrule
Definition: & A method where the LLM assesses its confidence and, if low, requests help from the user or a stronger LLM. \\
\cmidrule(l){1-2}
How We Use Them: & To improve the accuracy of the model when the LLM is unsure about its outputs. \\
\cmidrule(l){1-2}
Example: & The LLM says "I am not confident about constraint $C$" and asks the user to verify. \\
\addlinespace
\midrule[0.75pt]\midrule[0.75pt]

\multicolumn{2}{@{}l}{\textbf{Term: Parameters}} \\
\midrule
Definition: & Known quantities in the optimization problem, each with a symbol, shape, and definition. \\
\cmidrule(l){1-2}
How We Use Them: & Extracted from the problem description and used in clause formulations. \\
\cmidrule(l){1-2}
Example: & The cost per unit, denoted as $c$, is a parameter. \\
\addlinespace
\midrule[0.75pt]\midrule[0.75pt]

\multicolumn{2}{@{}l}{\textbf{Term: Variables}} \\
\midrule
Definition: & Unknown quantities to be determined in the optimization problem, each with a symbol, shape, definition, and type. \\
\cmidrule(l){1-2}
How We Use Them: & Defined during clause formulation and used in constraints and objectives. \\
\cmidrule(l){1-2}
Example: & The production quantity $x$ is a variable. \\
\addlinespace
\midrule[0.75pt]\midrule[0.75pt]

\multicolumn{2}{@{}l}{\textbf{Term: Background}} \\
\midrule
Definition: & A short string that explains the real-world context of the problem. \\
\cmidrule(l){1-2}
How We Use Them: & Included in every prompt to improve common sense reasoning. \\
\cmidrule(l){1-2}
Example: & "This problem involves optimizing factory production." \\
\addlinespace
\midrule[0.75pt]\midrule[0.75pt]

\multicolumn{2}{@{}l}{\textbf{Term: Error Correction}} \\
\midrule
Definition: & Techniques used to mitigate the impact of LLM hallucinations and correct errors. \\
\cmidrule(l){1-2}
How We Use Them: & Using reflective prompts and confidence-based user feedback to improve reliability. \\
\cmidrule(l){1-2}
Example: & Correcting a misidentified parameter by prompting "Is the value of $P$ known or not?" \\
\addlinespace
\midrule[0.75pt]\midrule[0.75pt]

\multicolumn{2}{@{}l}{\textbf{Term: Formulate Clauses}} \\
\midrule
Definition: & A process step where clauses are mathematically formulated, and variables and auxiliary constraints are defined. \\
\cmidrule(l){1-2}
How We Use Them: & To generate the mathematical representation of constraints and objectives. \\
\cmidrule(l){1-2}
Example: & Formulating "Total production must meet demand" as $x \geq d$. \\
\addlinespace
\midrule[0.75pt]\midrule[0.75pt]

\multicolumn{2}{@{}l}{\textbf{Term: Extract Parameters}} \\
\midrule
Definition: & A process step where parameters are extracted from the problem description. \\
\cmidrule(l){1-2}
How We Use Them: & To identify all known quantities needed for the optimization problem. \\
\cmidrule(l){1-2}
Example: & Extracting "demand" as a parameter from the description. \\
\addlinespace
\midrule[0.75pt]\midrule[0.75pt]

\multicolumn{2}{@{}l}{\textbf{Term: Correct Errors}} \\
\midrule
Definition: & A process step where errors in the parameters, clauses, and variables are corrected. \\
\cmidrule(l){1-2}
How We Use Them: & To ensure accuracy of the extracted and formulated components. \\
\cmidrule(l){1-2}
Example: & Correcting a parameter's shape if it was misidentified. \\
\addlinespace
\midrule[0.75pt]\midrule[0.75pt]

\multicolumn{2}{@{}l}{\textbf{Term: Extract Targets}} \\
\midrule
Definition: & A process step where constraints, objective, and connection graph are extracted. \\
\cmidrule(l){1-2}
How We Use Them: & To identify the clauses and their relationships to parameters and variables. \\
\cmidrule(l){1-2}
Example: & Extracting the constraint "Production capacity cannot exceed limit." \\
\addlinespace
\midrule[0.75pt]\midrule[0.75pt]

\multicolumn{2}{@{}l}{\textbf{Term: Code Targets}} \\
\midrule
Definition: & A process step where code snippets for all parameters, clauses, and variables are generated. \\
\cmidrule(l){1-2}
How We Use Them: & To assemble these snippets into a single runnable code file. \\
\cmidrule(l){1-2}
Example: & Generating code to define variables in a solver. \\
\addlinespace
\midrule[0.75pt]\midrule[0.75pt]

\multicolumn{2}{@{}l}{\textbf{Term: Assemble Code}} \\
\midrule
Definition: & A process step where all code snippets are combined into a single code file. \\
\cmidrule(l){1-2}
How We Use Them: & To produce executable code for solving the optimization problem. \\
\cmidrule(l){1-2}
Example: & Combining parameter definitions, variable declarations, and constraints into one script. \\
\addlinespace
\midrule[0.75pt]\midrule[0.75pt]

\multicolumn{2}{@{}l}{\textbf{Term: Debug}} \\
\midrule
Definition: & A process step where code is iteratively executed and errors are fixed until the desired results are achieved. \\
\cmidrule(l){1-2}
How We Use Them: & To correct syntax or runtime errors in the generated code. \\
\cmidrule(l){1-2}
Example: & Fixing a variable name mismatch that caused a runtime error. \\
\addlinespace
\bottomrule

\end{longtable}




\section{Reflexive Prompts} \label{app:reflex}
The reflective prompts used by OptiMUS-0.3 are as follows:

\paragraph{Parameter Extraction.} 
The model often confuses parameters with variables, misidentifies parameter shapes or misses some parameters. We use the following reflective prompt to correct these errors:

\begin{figure}
    \centering
\includegraphics{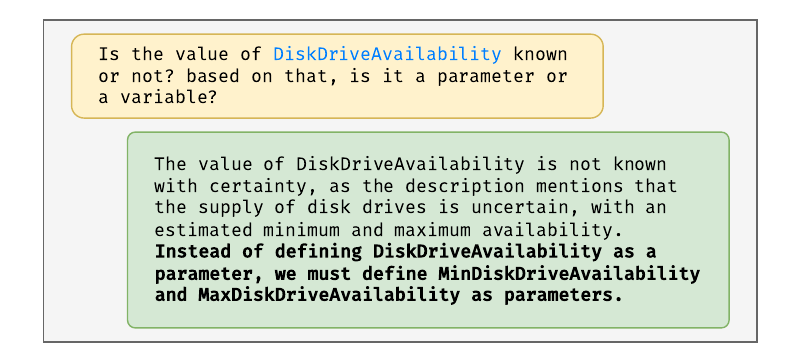}
    \caption{OptiMUS-0.3 can fix parameter identification errors when prompted \textit{``Is the value of $P$ known or not?''}}
    \label{fig:param-check}
\end{figure}

\begin{itemize}
    \item \textit{``Is the value of $P$ known or not? Based on that, is it a parameter or a variable?''} (see \cref{fig:param-check})
\end{itemize}

\paragraph{Constraint Extraction.}
During constraint extraction, 
the model sometimes extracts trivial or vague constraints 
and may regenerate a previously generated constraint in different language. 
We use the following reflective prompts to correct these errors: 
\begin{tcolorbox}[colback=OGray,  colframe=OGreen, coltitle=black, fonttitle=\sffamily\small, fontupper=\sffamily\small]

\begin{tcolorbox}[colback=OYellow,  colframe=OYellow]
  Does constraint $C$ need to be explicitly modeled in the mathematical formulation? (see  \cref{fig:constraint-explicitly})
\end{tcolorbox}

\begin{tcolorbox}[colback=OYellow,  colframe=OYellow]
    Here's the list of all constraints: $C_1, C_2, \ldots, C_k$. Are any of them redundant?
\end{tcolorbox}

\end{tcolorbox}



\begin{figure}
    \centering
\includegraphics{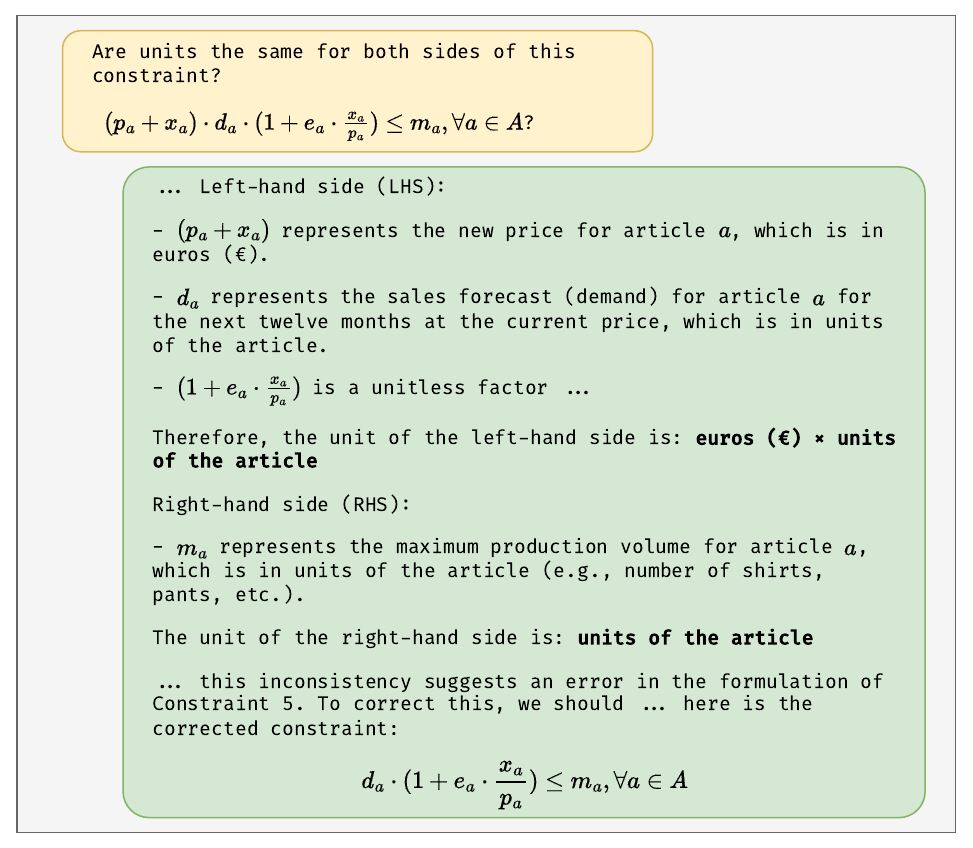}
    \caption{OptiMUS-0.3 can fix its constraint modeling errors when prompted ``\textit{Are units the same for both sides of $C$?}''}
    \label{fig:constraint-unit-check-appendix}
\end{figure}

\paragraph{Clause Modeling.}
Once the clauses are extracted, OptiMUS-0.3 models them, defining variables and auxiliary constraints if needed. 
Often, the model generates incorrect mathematical formulations for constraints. 
We use the following reflective prompts to detect and correct these errors:

\begin{tcolorbox}[colback=OGray,  colframe=OGreen, coltitle=black, fonttitle=\sffamily\small, fontupper=\sffamily\small]
	
\begin{tcolorbox}[colback=OYellow,  colframe=OGray]
  Are units the same for both sides of $C$? (see \cref{fig:constraint-unit-check})
\end{tcolorbox}

\begin{tcolorbox}[colback=OYellow,  colframe=OGray]
    Is $C$ a correct constraint in the context of MILP solvers? (see \cref{fig:constraint_context})
\end{tcolorbox}

\begin{tcolorbox}[colback=OYellow,  colframe=OGray]
    Does constraint $C$ involve any variables, or is it only defined over parameters?
\end{tcolorbox}

\end{tcolorbox}

\begin{figure}
    \centering
    \includegraphics{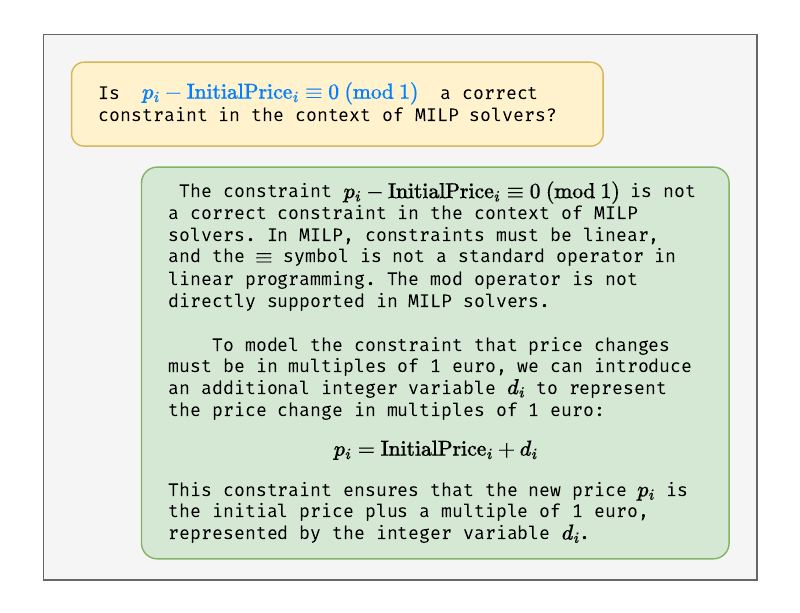}
    \caption{Through introspection, OptiMUS-0.3 can fix its errors in constraint modeling.}
    \label{fig:constraint_context}
\end{figure}

\begin{figure}
    \centering
\includegraphics{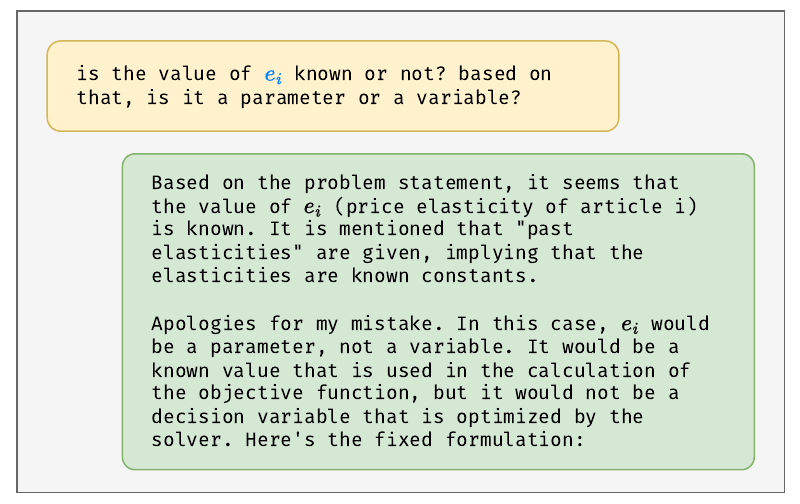}
    \caption{OptiMUS-0.3 can fix its parameter and variable identification errors when prompted \textit{``Is the value of $P$ known or not?''}}
    \label{fig:ec-ref-parameter}
\end{figure}


OptiMUS-0.3 uses similar reflective prompts to perform error checking for the optimization objective. For variables, we use the following prompt:


\begin{tcolorbox}[colback=OGray,  colframe=OGreen, coltitle=black, fonttitle=\sffamily\small, fontupper=\sffamily\small]
	
\begin{tcolorbox}[colback=OYellow,  colframe=OGray]
  Is the value of $V$ known or not? Based on that, is it a parameter or a variable? (see \cref{fig:ec-ref-parameter}) 
\end{tcolorbox}

\end{tcolorbox}




\begin{figure}[h]
    \centering
\includegraphics{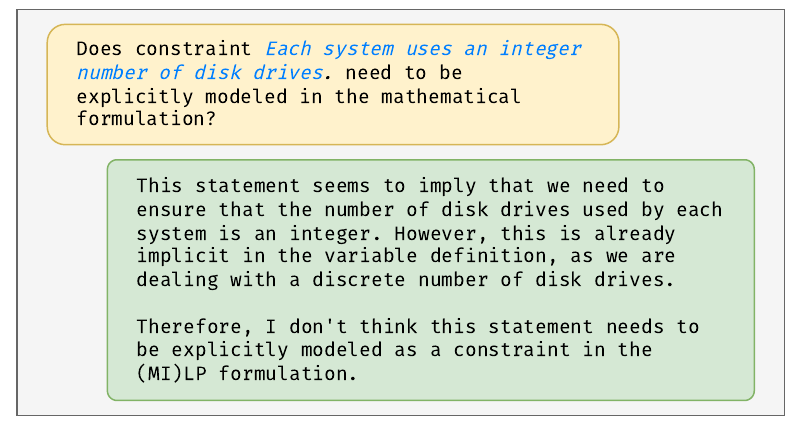}
    \caption{OptiMUS-0.3 can fix its parameter and variable identification errors when prompted \textit{``Is the value of $P$ known or not?''}}
    \label{fig:constraint-explicitly}
\end{figure}

\section{Additional details on Structure Detection} \label{app:structure_detection}

Table \ref{table:structure} shows the distribution of special problem structures in the NLP4LP dataset.

\begin{table}[h]
\centering
\caption{Distribution of problem structures in NLP4LP \label{table:structure}}
\begin{tabular}{cc}
\toprule
Structure & Number of instances \\
\midrule
 SOS         &  2				\\
 Indicator   & 6				\\
 Piecewise-linear & 17			\\
\bottomrule
\end{tabular}
\end{table}

\begin{figure}[ht]
    \centering
    
{
\begin{tcolorbox}[title=Example: Indicator Constraint, arc=0pt, colframe=OYellow, coltitle=black, colback=OGreen, 
fonttitle=\small, 
fontupper=\small]

{ 
  \begin{tcolorbox}[colback=OGray,  colframe=OYellow, coltitle=black, title=Definition]
    An indicator constraint over a binary variable \( z \) and a constraint \( C \)
    states that $z = 1 \Rightarrow C$.
  \end{tcolorbox}
}

{ 
  \begin{tcolorbox}[colback=OGray,  colframe=OYellow, coltitle=black, width=\textwidth, title=Relevant description]
    A retail company wants to open new stores in a set of \( K \) potential locations. A store at location \( k \) must have a minimum staff of \( n_k \) employees (to ensure adequate customer service) and a minimum inventory level of \( l_k \) units (to avoid stockout). The maximum inventory level for store \( k \) is \( u_k \), and there is a fixed cost associated with opening each store.
  \end{tcolorbox}
}

{ 
  \begin{tcolorbox}[colback=OGray,  colframe=OYellow, coltitle=black, title=Relevant Constraints]
    If a store at location \( k \) is open, it must have at least \( n_k \) employees. Moreover, let the inventory level of store \( k \) be \( y_k \).
  \end{tcolorbox}
}

{ 
  \begin{minipage}{0.42\textwidth}
    \begin{tcolorbox}[colback=OGray,  colframe=OYellow, coltitle=black, title=Formulations]
      {Original:}
      \vspace{-0.16cm}
      \begin{eqnarray*}
        x_k \in \{ 0, 1 \}, ~
        y_k \leq u_k x_k, \\
        z_k \geq l_k - M (1 - x_k)
      \end{eqnarray*}
      {OptiMUS:}
      \vspace{-0.16cm}
      \begin{eqnarray*}
        x_k & \Rightarrow & y_k \leq u_k \\
        x_k & \Rightarrow & z_k \geq l_k
      \end{eqnarray*}
    \end{tcolorbox}
  \end{minipage}
  \hfill
  \begin{minipage}{0.55\textwidth}
    \begin{tcolorbox}[colback=white,  colframe=OYellow, coltitle=black, title=Performance plot]
      \centering
      \includegraphics[scale=0.24]{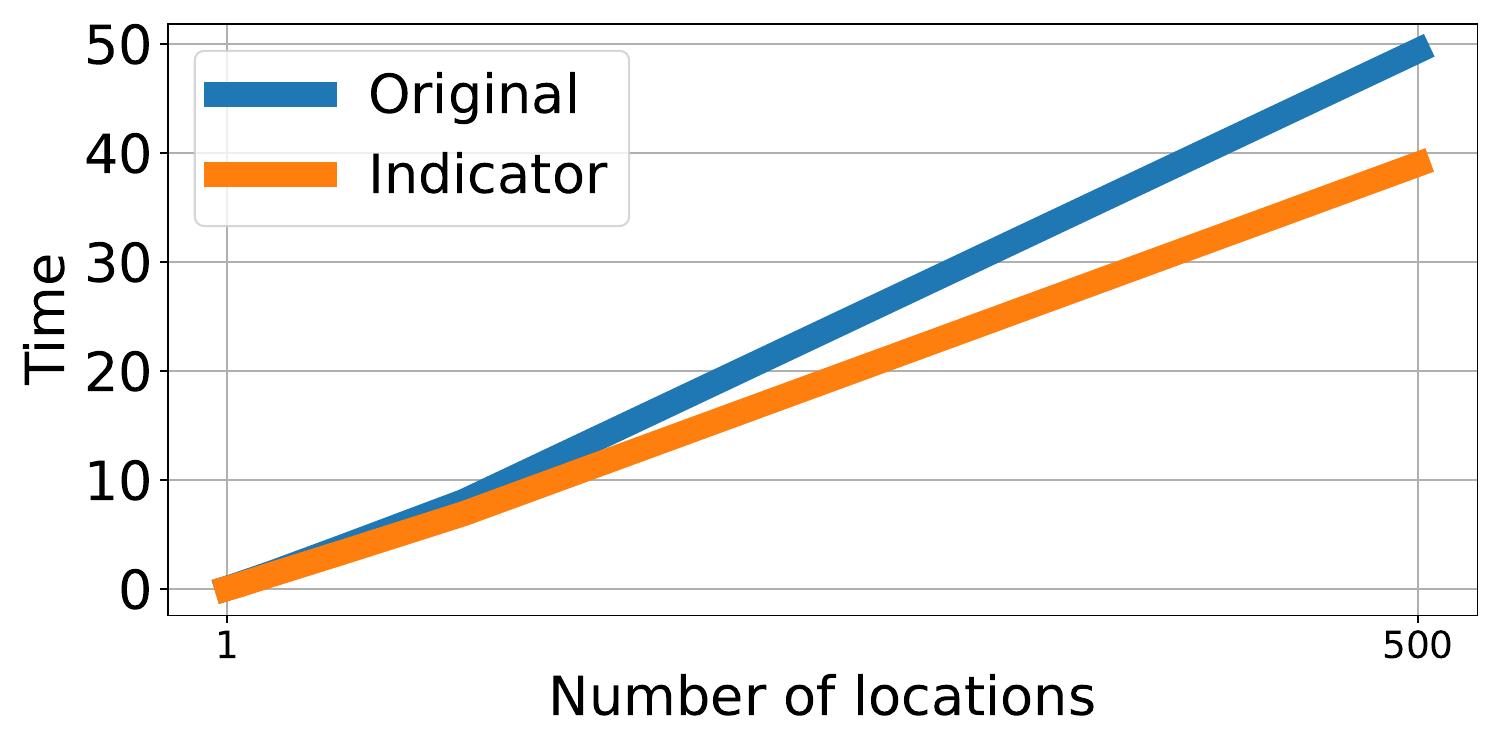}
    \end{tcolorbox}
  \end{minipage}
}
\end{tcolorbox}

}

\end{figure}

\begin{figure}[ht]
    \centering
{ 
\begin{tcolorbox}[title=Example: Special Ordered Set, colback=OGreen,  colframe=OYellow, coltitle=black, 
fonttitle=\small, 
fontupper=\small]

{ 
\begin{tcolorbox}[colback=OGray,  colframe=OYellow, coltitle=black, title=Definition]
  Given a set of variables $\mathcal{X}= \{ x_1, \ldots, x_n \}$,
  \begin{itemize}
    \item $\{ x_1, \ldots, x_n \} \in \tmop{SOS}_1$ if at most one element of
    $\mathcal{X}$ can be nonzero
    
    \item $\{ x_1, \ldots, x_n \} \in \tmop{SOS}_2$ if at most two elements of
    $\mathcal{X}$ can be nonzero
  \end{itemize}
\end{tcolorbox}
}

{
{ 
  \begin{tcolorbox}[colback=OGray,  colframe=OYellow, coltitle=black, title=Relevant Description]
    An airline needs to assign $N$ crew members to a set of $K$ flights for a day. Each crew member can only be assigned to at most one flight at a time, and there must be a minimum rest period of 2 hours between consecutive flights to comply with aviation regulations and ensure crew well-being. Each crew member has a maximum number of flying hours per day, set at 8 hours, to comply with aviation regulations and prevent fatigue. There are minimum and maximum layover times between flights: minimum layover of 2 hours (to allow for crew rest and flight preparations) and maximum layover of 6 hours (to optimize crew utilization and reduce idle time). Assigning a crew member to a specific flight has some cost.
  \end{tcolorbox}
}

{ 
  \begin{tcolorbox}[colback=OGray,  colframe=OYellow, coltitle=black, title=Relevant Constraints,
  fontupper= 
  ]
    Each crew member can work on only one flight. Let $x_{i j} \in \{ 0, 1 \}$ denote whether crew $i$ works on flight $j$ and $T_{i j}$ denote the working time of crew $i$ on flight $j$.
  \end{tcolorbox}
}

{ 
  \begin{minipage}{0.40\textwidth}
  \begin{tcolorbox}[colback=OGray,  colframe=OYellow, coltitle=black, title=Formulations]
{Original:}
\begin{align*}
	\textstyle \sum_{j = 1}^K x_{i j} \leq 1, \quad &\forall i \in [N], \\
	T_{i j} \leq  24 x_{i j}, \quad &\forall i \in [N], j \in [K]
\end{align*}
{OptiMUS:}

\[\{ T_{i 1}, \ldots, T_{i K} \} \in \tmop{SOS}_1\]\
\end{tcolorbox}
\end{minipage}
\hspace{2pt}
\begin{minipage}{0.55\textwidth}
  \begin{tcolorbox}[colback=white,  colframe=OYellow, coltitle=black,title=Performance plot]
  \centering
\includegraphics[scale=0.27]{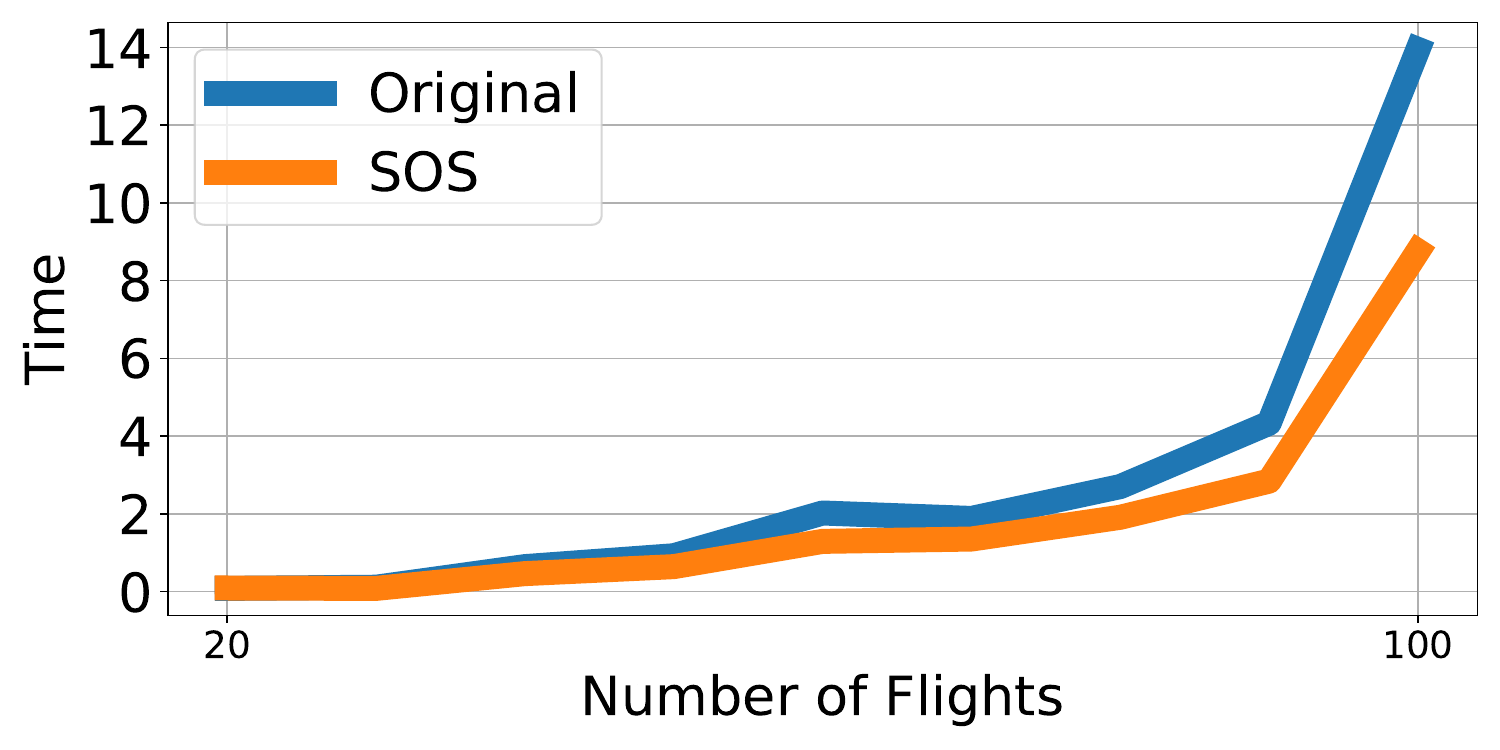}
\end{tcolorbox}
\end{minipage}
}
  }
\end{tcolorbox}
}

\end{figure}

\section{SCUC problem} \label{app:scuc}

\textbf{Description.} The Security Constrained Unit Commitment (SCUC) is a
critical optimization problem in the operation of electrical power systems.
Its goal is to schedule generation units in a cost-effective manner
while ensuring reliable operation over a specified time period (typically 24
hours), considering the system's physical and operational constraints. This appendix presents
a detailed description of the SCUC problem, including its objectives and
constraints:
\noindent
\textbf{Objective.}
The objective of the SCUC problem is to minimize the total operational
cost of the power system. This cost generally includes:
\begin{itemize}[leftmargin=10pt]
\item Fuel Cost: The cost associated with consuming fuel to generate electricity.
\item Start-up and Shut-down Cost: Cost incurred from starting up or shutting down
generation units.
\item Emission Cost: Cost related to the emissions produced by the generating
units, if applicable.
\end{itemize}
\textbf{Constraints.}
The SCUC must satisfy several constraints to ensure the safe and reliable
operation of the power system:
\begin{itemize}[leftmargin=10pt]
    \item Power Balance Constraint: Ensures that the total power generation meets the
total demand plus system losses at every interval. Mathematically, this is
expressed as the sum of the outputs of all online generators being equal to
the sum of the demand and the transmission losses.
\item Minimum and Maximum Output Limits: Each generator has a minimum and maximum
generation capacity when it is online. The SCUC ensures that the output of
each unit stays within these bounds.
\item Ramp-up and Ramp-down Limits: These limits specify the maximum rate at which a
generator can increase or decrease its output. They are critical for handling
load changes throughout the day.
\item Unit Start-up and Shut-down Constraints: These constraints handle the
logistics of turning units on or off. Start-up constraints may include minimum
down times (the minimum time a unit must remain off before it can be
restarted) and minimum up times (the minimum time a unit must remain on once
started).
\item Minimum Up and Down Time Constraints: Ensures that once a generator is turned
on, it stays on for at least its minimum up time, and similarly, once turned
off, it remains off for at least its minimum down time. These constraints are
crucial for the mechanical integrity of the generation units.
\item Reserve Requirements: The system operator must ensure that sufficient spinning
and non-spinning reserves are available. These reserves are needed to handle
sudden increases in demand or unexpected generator failures.
\end{itemize}

\section{Web Application}
\label{app:webapp}

As discussed in \ref{intro}, the OptiMUS project aims to 1) help optimization experts develop and maintain their models with ease and 2) empower domain experts to make better decisions faster and reduce costs.
To further these goals, we designed and built a user-centric web app with an intuitive interface that enables users to interact seamlessly with the system and leverage the advantages of large language models (LLMs).

We had the following key characteristics in mind when designing the web app:

\begin{itemize}
    \item \textbf{User-centered design:} The design centers around the user's modeling flow rather than requiring users to adapt their workflow to the system.
    \item \textbf{Observability and intervention:} Given that LLM-based systems often are prone to errors, user supervision is critical. The interface is designed to facilitate this in every step.
\end{itemize}

The process includes several key steps:

    \begin{figure}
    \centering
    \includegraphics[width=0.8\textwidth]{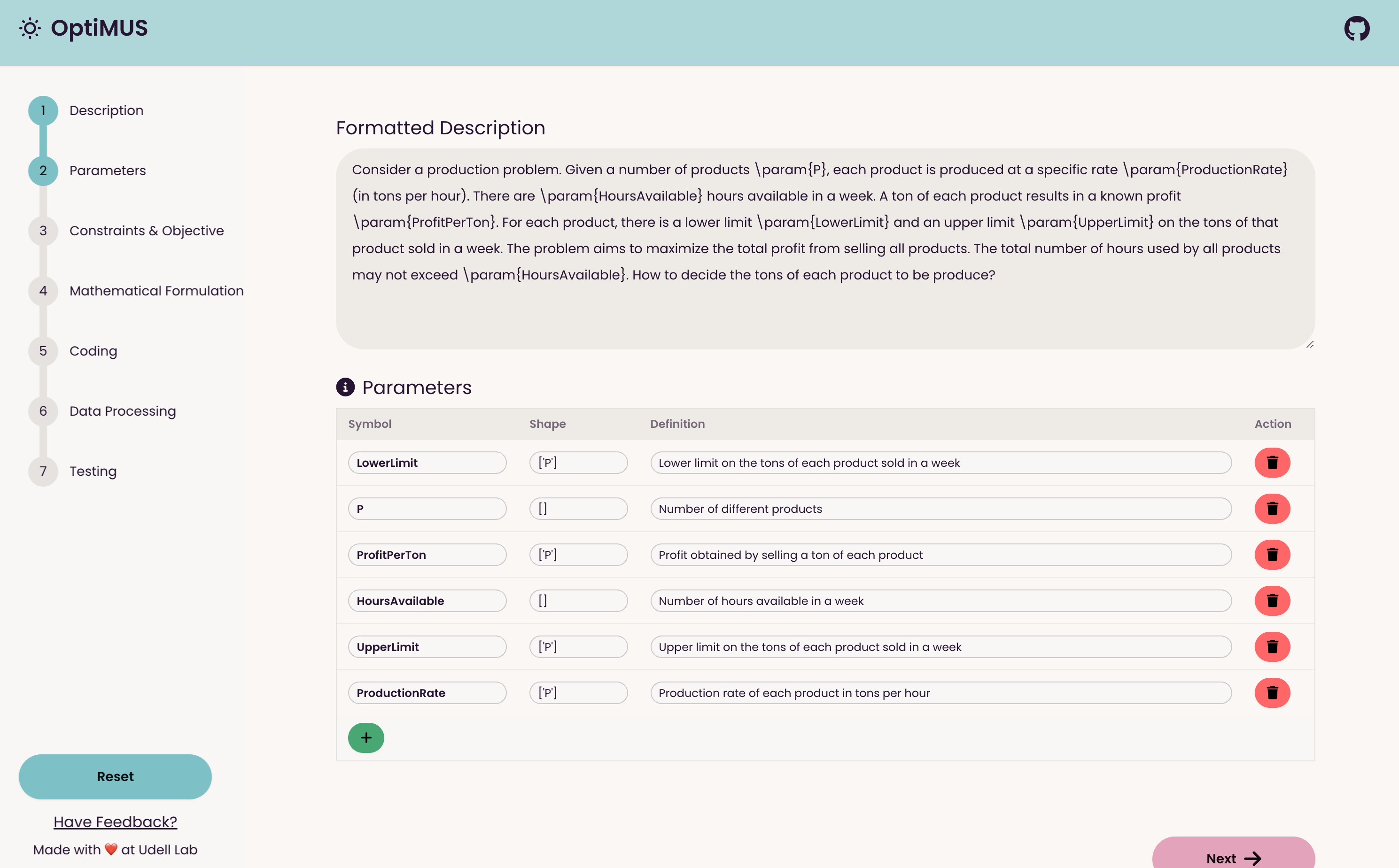}
    \caption{Parameter Identification: The parameters are automatically extracted, and users can add, remove, or modify them as needed.}
    \label{fig:parameters}
    \end{figure}
    
    \begin{figure}
    \centering
    \includegraphics[width=0.8\textwidth]{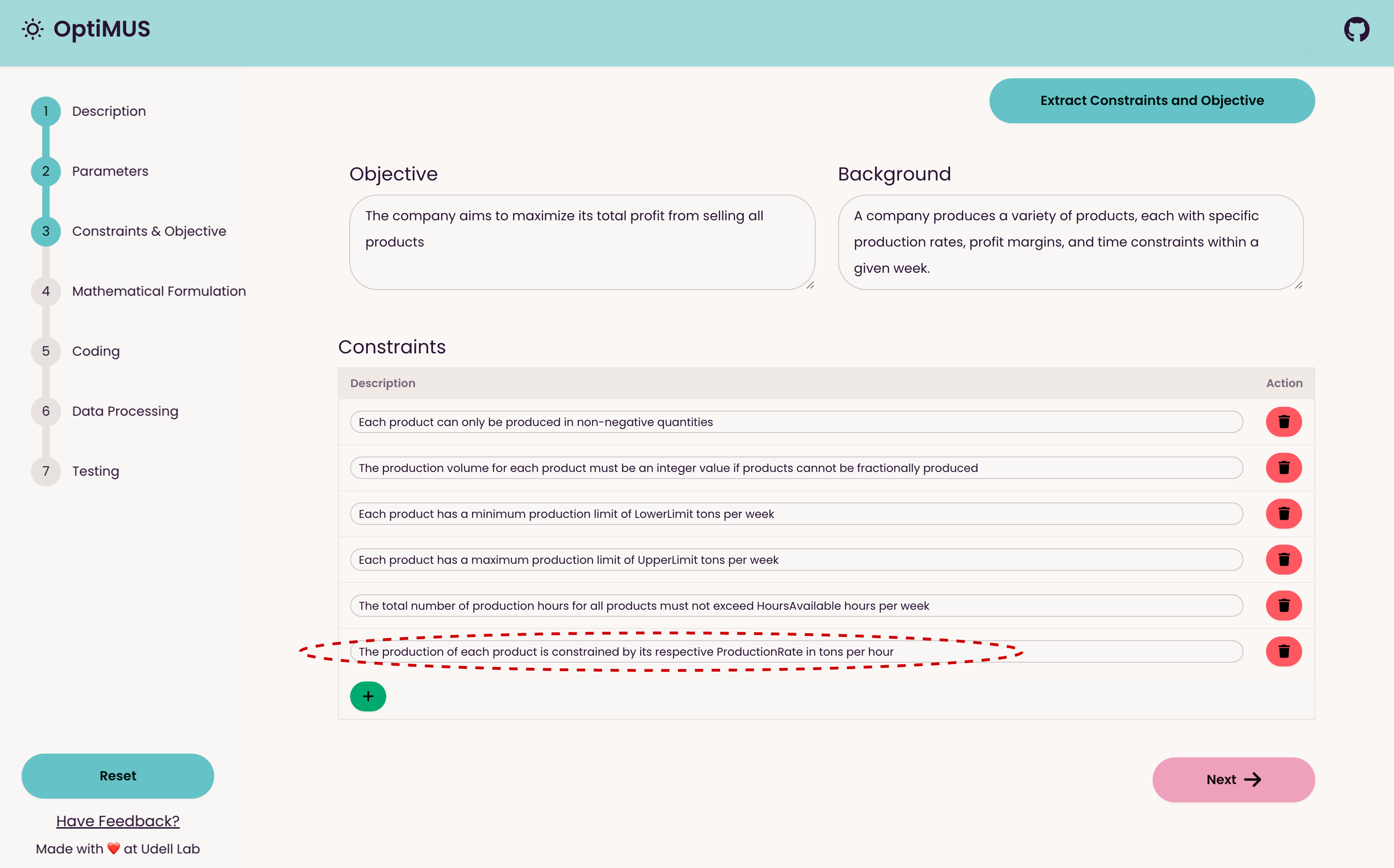}
    \caption{Clause Detection: In this example, one identified constraint is incorrectly marked as a constraint (The production limit is automatically enforced and does not need to be an explicit constraint). The interface allows the user to easily remove it.}
    \label{fig:clause_detection}
    \end{figure}
    
    \begin{figure}
    \centering
    \includegraphics[width=0.8\textwidth]{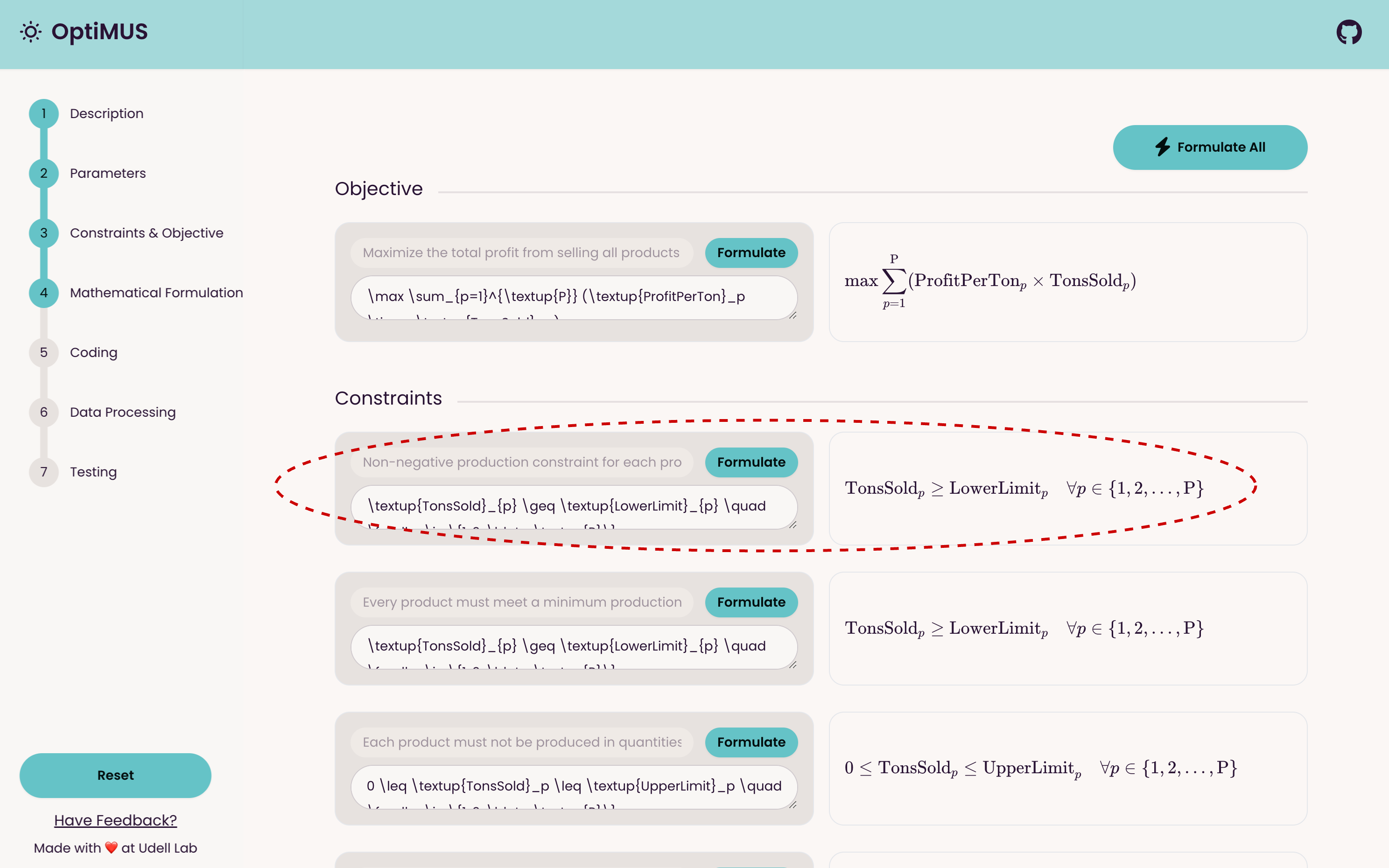}
    \caption{Clause Formulation: An error is identified where one constraint is incorrect, allowing the user to correct it.}
    \label{fig:clause_formulation}
    \end{figure}
    
    \begin{figure}
    \centering
    \includegraphics[width=0.8\textwidth]{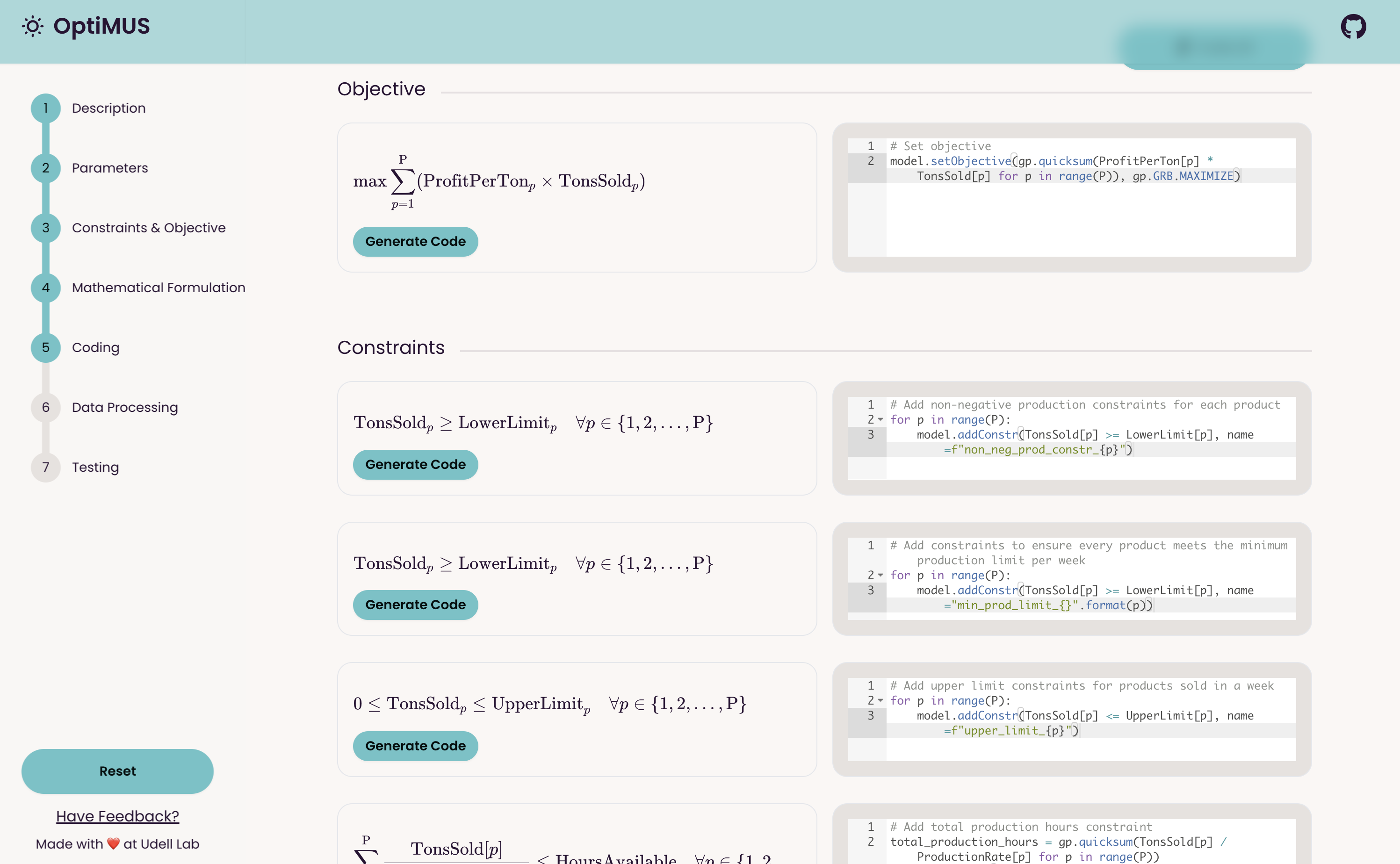}
    \caption{Code Generation for each clause based on the formulated constraints and objectives.}
    \label{fig:code_generation}
    \end{figure}

 \begin{figure}
    \centering
    \includegraphics[width=0.8\textwidth]{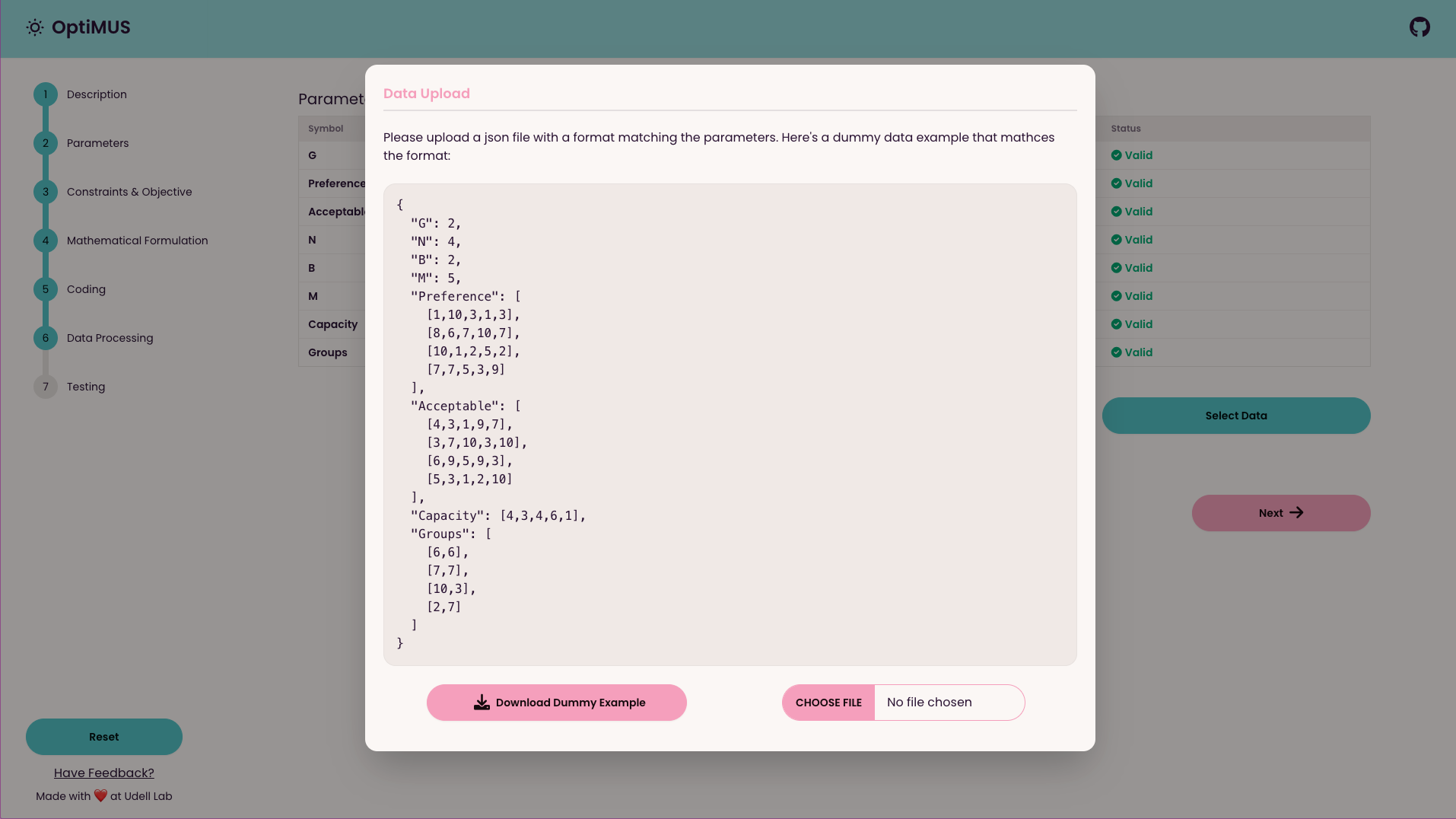}
    \caption{Data File Structure: Inferred from the extracted parameters, with options for data upload or random generation.}
    \label{fig:data_structure}
    \end{figure}

 \begin{figure}
    \centering
    \includegraphics[width=0.8\textwidth]{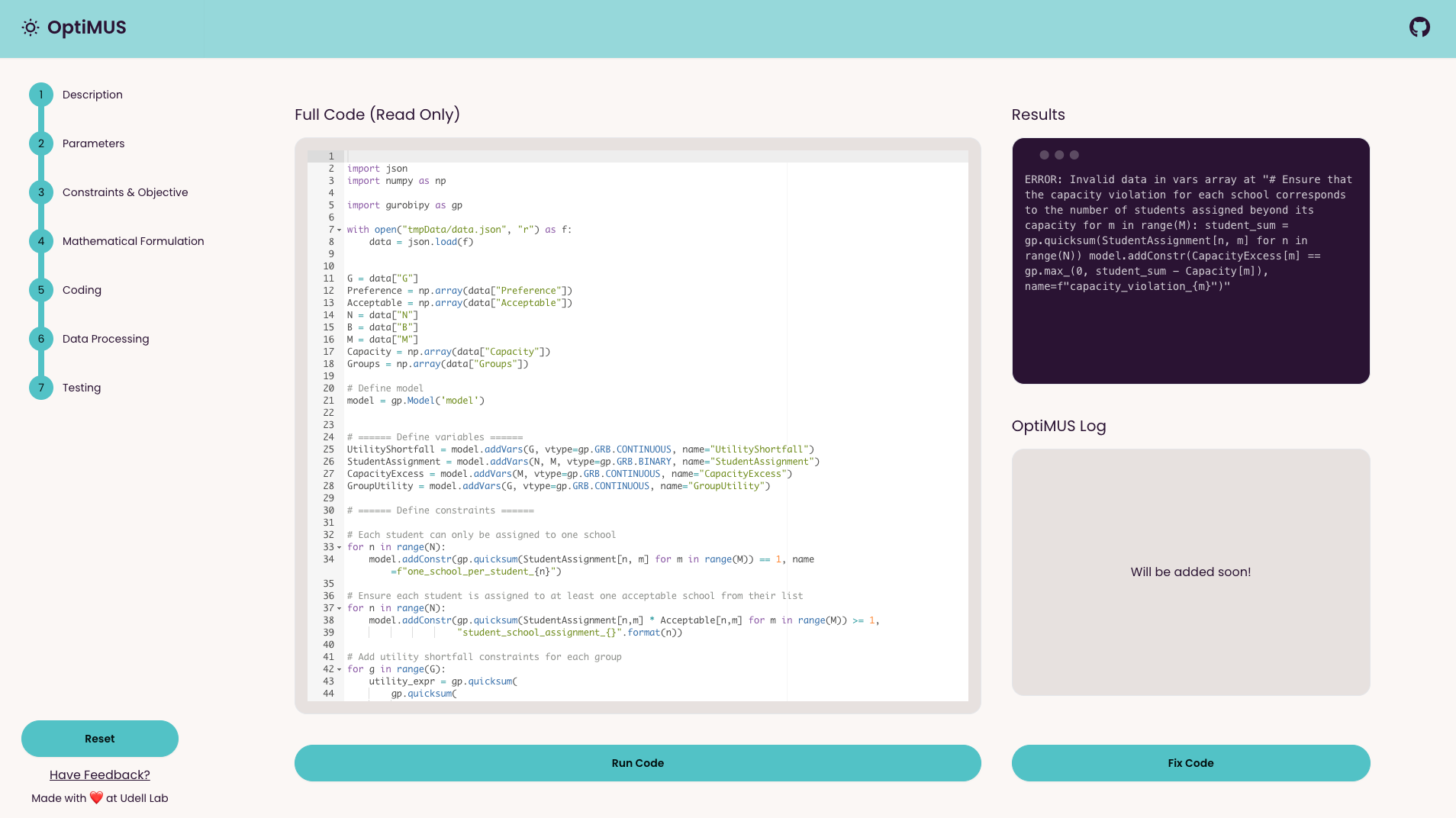}
    \caption{Testing Phase: The code is debugged and revised as errors are detected.}
    \label{fig:testing_phase}
    \end{figure}
    
    \begin{figure}
    \centering
    \includegraphics[width=0.8\textwidth]{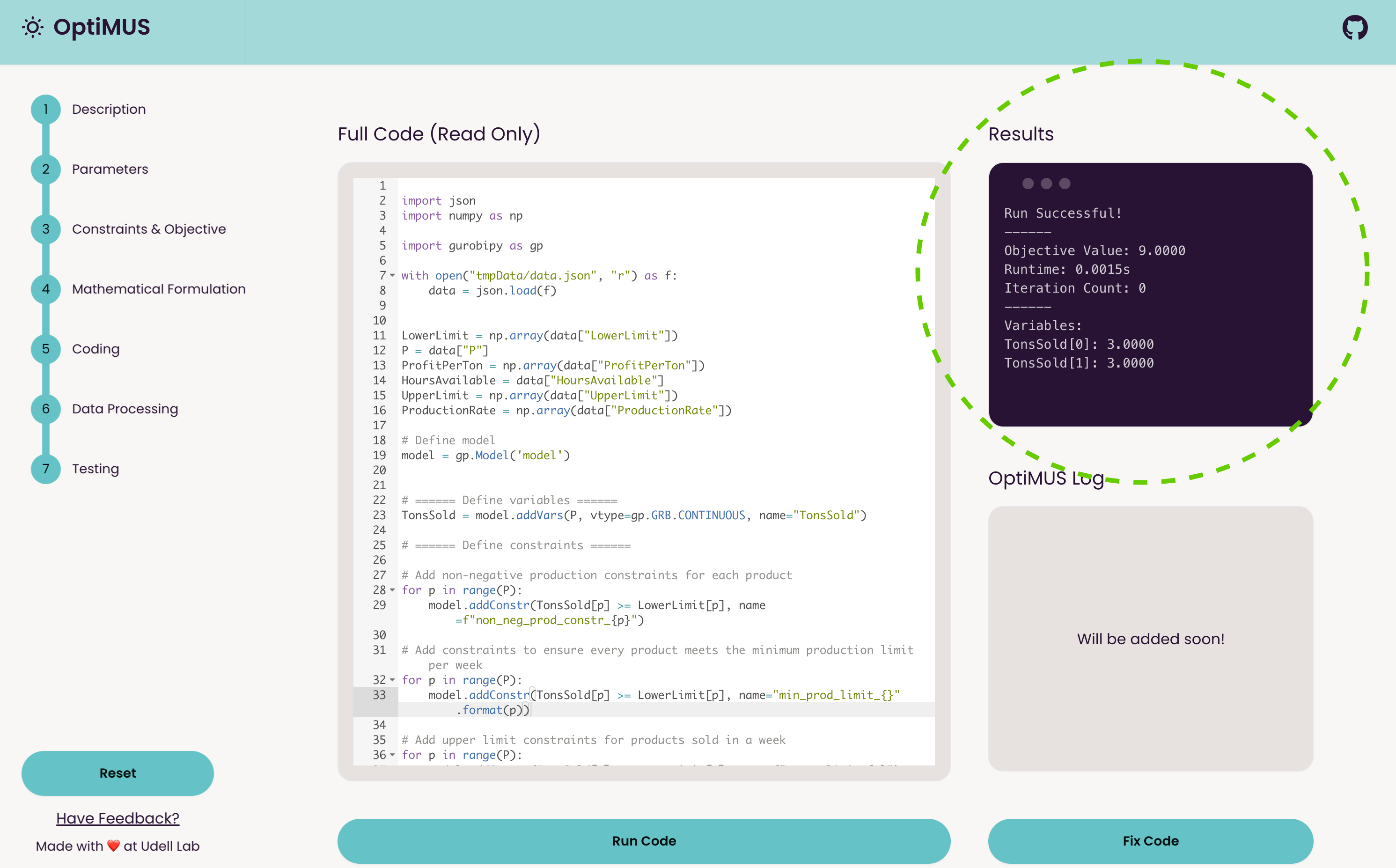}
    \caption{Final Output: After debugging, the correct code is executed and the solution is displayed.}
    \label{fig:final_output}
    \end{figure}

\begin{enumerate}
    \item \textbf{Parameter Extraction:} Users can input text in any format, and the model automatically extracts parameters. Users have the flexibility to edit and update parameters if necessary (Fig.~\ref{fig:parameters}).

    \item \textbf{Clause Detection:} The model detects the objective and constraints given the provided information (Fig.~\ref{fig:clause_detection}).

    \item \textbf{Clause Formulation:} Each clause is formulated by the model (Fig.~\ref{fig:clause_formulation}).

    \item \textbf{Code Generation:} The model generates code for each formulated clause (Fig.~\ref{fig:code_generation}).

    \item \textbf{Data File Structure Inference:} Based on the parameter information, a data file structure is inferred. Users can upload data in this format or use randomly generated data to continue (Fig.~\ref{fig:data_structure}).

    \item \textbf{Code Editing and Testing:} Users can edit the generated code, run it on the dataset, and debug it with assistance from the LLM (Figs.~\ref{fig:testing_phase} and \ref{fig:final_output}).
    
\end{enumerate}

By allowing users to observe and provide input throughout each step, the web app brings significant speed and convenience to the modeling process while minimizing error risks.

\newpage
\section{Dataset Figures}

\cref{fig:ds-types} and \cref{fig:ds-domain} represent the number of instances in the dataset per industry sectors and operational areas. Resource allocation is the most common area and manufacturing is the most common sector. Figure \ref{fig:ds-label-count} shows the distribution of instances by the number of associated labels with the problem. Most instances have around 4 labels, with some having as few as 2 and as many as 8

For plagiarism tests we use Papers Owl online tool \cite{papersowl2024plagiarismchecker}. See \cref{fig:plagiarism} for an example and \cref{fig:plagiarism_histogram} for a histogram of originality scores for all instances.

\begin{figure}[htbp]
\centering
\begin{tikzpicture}
\begin{axis}[
    ybar,
    symbolic x coords={2, 3, 4, 5, 6, 7, 8},
    xtick=data,
    ylabel={Count},
    title={Histogram of labels},
    width=0.5\textwidth
]
\addplot coordinates {(2, 6) (3, 52) (4, 144) (5, 97) (6, 37) (7, 12) (8, 1)};
\end{axis}
\end{tikzpicture}
\caption{Most problems have around 4 labels.}
\label{fig:ds-label-count}
\end{figure}
\label{app:dataset}
\begin{figure}[t]
\centering
\begin{tikzpicture}
\begin{axis}[
    ybar,
    symbolic x coords={Resource Allocation, Production Planning, Scheduling, Blending, Workforce Management, Selection, Sizing, Routing, Inventory Management, Packaging, Pricing, Balancing, Portfolio Optimization, Network Flow, Covering, Cutting, Fitting, Layout},
    xtick=data,
    x tick label style={rotate=45, anchor=east},
    ylabel={Count},
    title={Operational Areas},
    nodes near coords,
    width=0.90\textwidth
]
\addplot coordinates {(Resource Allocation, 324) (Production Planning, 179) (Scheduling, 80) (Blending, 55) (Workforce Management, 37) (Selection, 36) (Sizing, 33) (Routing, 28) (Inventory Management, 17) (Packaging, 15) (Pricing, 13) (Balancing, 13) (Portfolio Optimization, 10) (Network Flow, 8) (Covering, 4) (Cutting, 3) (Fitting, 3) (Layout, 2)};
\end{axis}
\end{tikzpicture}
\caption{NLP4LP covers common optimization problem types}
\label{fig:ds-types}
\end{figure}

\begin{figure}[t]
\centering
\begin{tikzpicture}
\begin{axis}[
    ybar,
    symbolic x coords={Manufacturing and Production, Supply Chain Management, Food and Beverage, Transportation and Logistics, Healthcare and Medical, Retail and E-commerce, Environmental and Sustainability, Agriculture and Forestry, Science and Research, Energy and Power Systems, Finance and Banking, Sports and Entertainment, Government and Public Sector, Education, Human Resources, Telecommunications, Marketing and Media, Aerospace and Defense},
    xtick=data,
    x tick label style={rotate=45, anchor=east},
    ylabel={Count},
    title={Industry Sectors},
    nodes near coords,
    width=0.90\textwidth
]
\addplot coordinates {(Manufacturing and Production, 142) (Supply Chain Management, 102) (Food and Beverage, 78) (Transportation and Logistics, 72) (Healthcare and Medical, 68) (Retail and E-commerce, 53) (Environmental and Sustainability, 30) (Agriculture and Forestry, 29) (Science and Research, 19) (Energy and Power Systems, 15) (Finance and Banking, 15) (Sports and Entertainment, 15) (Government and Public Sector, 13) (Education, 12) (Human Resources, 12) (Telecommunications, 3) (Marketing and Media, 3) (Aerospace and Defense, 2)};
\end{axis}
\end{tikzpicture}
\caption{NLP4LP covers important optimization problem domains}
\label{fig:ds-domain}
\end{figure}

\begin{figure}[t]
\centering
\includegraphics[width=\textwidth]{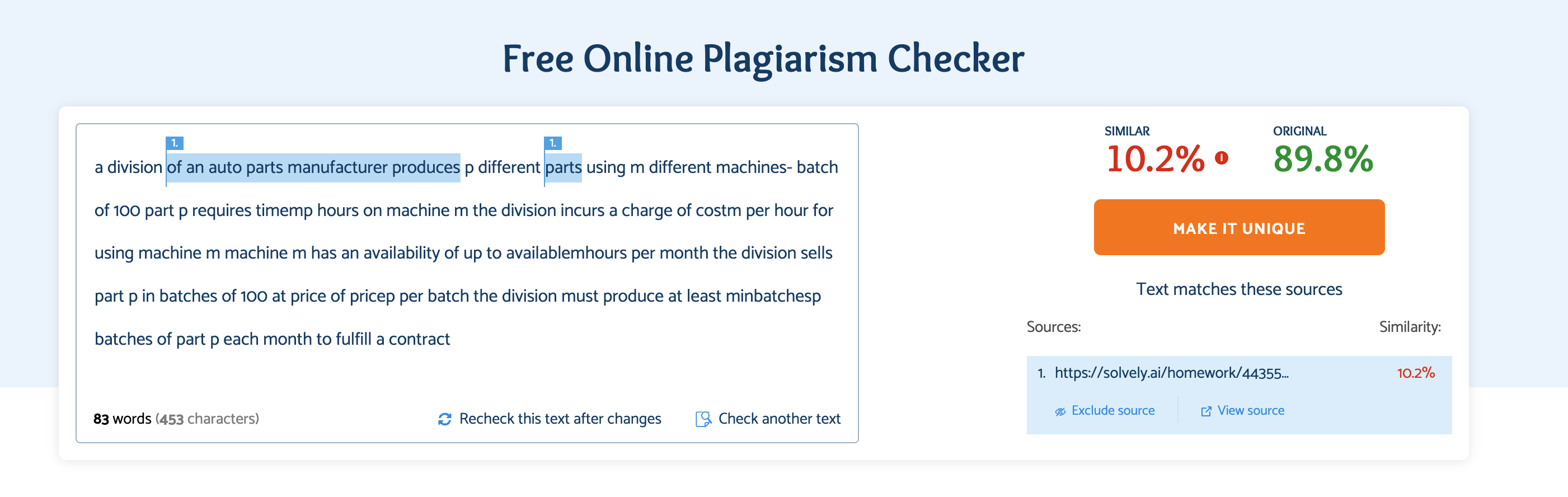}
\caption{We used plagiarism detection tools to ensure that instances are novel and publicly available on the internet. Most of the similarities detected are at word level (e.g. from a news article about manufacturing, or from a different optimization problem in the same domain).}
\label{fig:plagiarism}
\end{figure}

\begin{figure}
\begin{tikzpicture}
\begin{axis}[
    ybar,
    xlabel={Originality Score},
    ylabel={Instance Count},
    ymin=0,
    xmin=-9, xmax=109,
    ylabel style={rotate=-90},
    width=0.90\textwidth,
    height=8cm,
    bar width=24pt,
    enlarge x limits={abs=1pt},
]

\addplot+[ybar, fill=orange] coordinates {
    (30, 2) 
    (40, 3) 
    (50, 4) 
    (60, 6) 
    (70, 2) 
    (80, 2) 
    (90, 6) 
    (100, 34) 
};
\label{fig:plagiarism_histogram}
\end{axis}
\end{tikzpicture}
\caption{Originality of the NLP4LP dataset. The instances in our dataset represent a really high originality score overall, minimizing the chances of them being using in internet-scraped training corpus used for training LLMs.}
\end{figure}

\newpage
\section{Inference Settings}
\label{app:inference_settings}

\caledit{For reproducibility, we report the exact inference settings used for all LLM calls in our experiments. No inference parameters were set explicitly in the OptiMUS code; all values listed below reflect the API defaults for each model at the time of our experiments.}

\begin{table}[h]
\centering
\caption{\caledit{Inference settings for all LLM calls in OptiMUS experiments.}}
\label{table:inference_settings}
\small
\begin{tabular}{p{3.2cm}p{3.8cm}ccc}
\toprule
Use & Model (version) & Temperature & Top-$p$ & Max Output Tokens \\
\midrule
Main pipeline & GPT-4o (gpt-4o-2024-11-20) & 1.0 & 1.0 & 16,384 \\
Parameter extraction & o1 & 1.0$^\dagger$ & 1.0$^\dagger$ & 100,000 \\
OptiMUS+o3 experiments & o3 & N/A$^\ddagger$ & N/A$^\ddagger$ & 100,000 \\
Open-source experiments & LLaMA-3.1-70B-Instruct (llama3-70b-8192) & 1.0 & 1.0 & 8,192 \\
\bottomrule
\end{tabular}
\end{table}

\caledit{\noindent $^\dagger$ Temperature and top-$p$ are fixed at 1.0 for o1 and cannot be adjusted via the API.\\
\noindent $^\ddagger$ Temperature and top-$p$ are not supported parameters for o3; reasoning depth is controlled internally by the model.}

\end{APPENDIX}
\end{document}